\pdfoutput=1
\documentclass[11pt]{article}
\usepackage[final]{acl}
\usepackage{times}
\usepackage{latexsym}
\usepackage[T1]{fontenc}
\usepackage[utf8]{inputenc}
\usepackage{microtype}
\usepackage{inconsolata}
\usepackage{graphicx}
\usepackage{booktabs}
\usepackage{multirow}
\usepackage{amsmath}
\usepackage{xcolor}
\usepackage{tabularx}
\usepackage{enumitem}
\usepackage{tcolorbox}
\usepackage{pifont}
\usepackage{fontawesome5}
\usepackage{placeins}
\newcommand{\yes}{\textcolor{black!75}{\ding{51}}}
\newcommand{\no}{\textcolor{black!75}{\ding{55}}}

\definecolor{takeawayback}{RGB}{243,247,252}
\definecolor{takeawaybulb}{RGB}{226,152,18}
\newtcolorbox{takeawaybox}{colback=takeawayback,colframe=black!30,boxrule=0.4pt,arc=1.5pt,left=4pt,right=4pt,
  top=3pt,bottom=3pt,boxsep=0pt,before skip=9pt,after skip=9pt}
\newcommand{\takeaway}[1]{\begin{takeawaybox}\small{\color{takeawaybulb}\faLightbulb}\hspace{0.3em}\textbf{Takeaway.}
#1\end{takeawaybox}}
\newcommand{\jev}{Jev}
\newcommand{\luna}{GPT-6 Luna}
\newcommand{\lunad}{GPT-6 Luna Decisions}
\newcommand{\kev}{Kev-27B}
\newcommand{\qwen}{Qwen3.8-27B}
\newcommand{\llama}{Llama-3.3-70B}

\title{Typed Decision Models on Tabular Data}

\author{Baowen Zhang\textsuperscript{1},
  Hangting Ye\textsuperscript{2}\thanks{\hspace{0.3em}Corresponding author.},
  Hongyuan Zha\textsuperscript{3} \\
  \textsuperscript{1}University of Wisconsin--Madison \\
  \textsuperscript{2}School of Artificial Intelligence, Jilin University \\
  \textsuperscript{3}School of Data Science, Chinese University of Hong Kong, Shenzhen \\
  \texttt{bzhang599@wisc.edu}, \texttt{yeht22@mails.jlu.edu.cn}, \texttt{zhahy@cuhk.edu.cn}}

\begin{document}
\maketitle

\begin{abstract}
Typed decision models are language models that answer a question with a fixed set of possible answers by returning a
probability for each answer instead of generating text. We study them on tabular data and compare four decision models
with three large language models (LLMs) on row classification, fact verification and anomaly detection. Every request
is sent in its original form and under 21 types of rewrite that keep its content, such as renaming the options or
writing the table in another format. The decision models perform as well as the LLMs at a lower cost. Three of them are also far more robust to how the question is formulated. No model we test is robust to how the table is organized. For every model, at least a third of the row decisions change under some rewrite of the table or the question. The decisions to which a decision
model gives low probability are the ones most likely to change. Decision models are a practical choice for tabular decisions, but their robustness remains an open problem.
\end{abstract}

\section{Introduction}\label{sec:intro}

Tabular data, in which each row describes a sample and each column one of its heterogeneous features, is one of the most common forms of
data in fields such as healthcare, finance and network security \citep{borisov2022deep,fang2024llmtabular}.
On such data, gradient-boosted
trees \citep{chen2016xgboost,ke2017lightgbm,prokhorenkova2018catboost} still outperform most deep networks designed
for tables \citep{grinsztajn2022tree,borisov2022deep}, and tabular foundation models do best on small tables
\citep{hollmann2025tabpfn,priorlabs2026tabpfn35}.

\begin{table*}[t]
\centering\footnotesize
\setlength{\tabcolsep}{5pt}
\def\yes{\textcolor[RGB]{0,140,70}{\ding{51}}}
\def\no{\textcolor[RGB]{200,40,40}{\ding{55}}}
\def\partly{\textcolor[RGB]{210,130,0}{\textbf{Partly.}}}
\begin{tabularx}{\textwidth}{@{}>{\raggedright\arraybackslash}p{4.6cm}>{\raggedright\arraybackslash}X
>{\raggedright\arraybackslash}X>{\raggedright\arraybackslash}X@{}}
\toprule
& \multicolumn{2}{c}{Decision models} & \\
\cmidrule(lr){2-3}
Research question & \jev{}, Clef, \kev{} & \lunad{} & LLMs \\
\midrule
\textbf{RQ1} Are decision models practical? (\S\ref{sec:rq1}) & \yes{} As accurate as LLMs and best on two of three tasks, at far lower cost & \yes{} As accurate as \luna{} and cheaper on rows and samples & \yes{} Best on row
classification, but a decoding pass per decision \\
\addlinespace[2pt]
\textbf{RQ2} Are they robust to the formulation of the question? (\S\ref{sec:rq2}) & \yes{} Option names and order
barely matter & \no{} Reversing the options changes up to 23\% & \no{} Option names and order change up to 30\% \\
\addlinespace[2pt]
\textbf{RQ3} Are they robust to how the table is organized? (\S\ref{sec:rq3}) & \no{} Format and row and column order
change many decisions & \no{} Likewise & \no{} Likewise \\
\addlinespace[2pt]
\textbf{RQ4} Does their confidence indicate which decisions are sensitive to rewrites? (\S\ref{sec:rq4}) & \yes{} Low
confidence marks the decisions a rewrite changes & \partly{} As well as an LLM's best readout & \partly{} Only with ten
draws or token probabilities \\
\bottomrule
\end{tabularx}
\caption{Our four research questions and their answers for the decision models and the LLMs.}
\label{tab:glance}
\end{table*}

Large language models (LLMs) have opened a new route for tabular data \citep{fang2024llmtabular}.
LLM-based methods first serialize a row or a table into text, for example as feature--value pairs or as a Markdown table \citep{hegselmann2023tabllm,sui2024tablemeetsllm}. The serialized row is then classified by prompting an LLM with a description of the task and a few in-context
examples, or by fine-tuning it \citep{dinh2022lift,hegselmann2023tabllm}. The LLM can also turn the labelled rows
into features and rules for a conventional model \citep{han2024featllm,ye2025delta}. This works well even when only a few labelled rows are available
\citep{hegselmann2023tabllm}. However, LLMs are not built for decisions. The LLM-generated output has to be mapped to
one of the options, and every decision needs a full decoding process, which can be expensive when a large number of decisions need to be made. Their answers can also change with the order of the columns, the format of the rows or the wording
of the question \citep{sui2024tablemeetsllm,liu2024rethinking,sclar2024quantifying}.

Recently, a new class of language models has been built for such decisions: \emph{typed decision models} (decision models for short). TypeSafe released the first of them, \jev{} \citep{typesafe2026jev}. \jev{} takes a sample or a table together with one or more questions, each with a fixed set of possible answers
such as the classes of a task. Instead of generating text, it returns a probability for every answer of every
question in a single request. TypeSafe trains it with Reinforcement Learning for Calibrated Decisions (RLCD) so that
these probabilities match outcomes, and reports that its confidence is calibrated and that similar inputs receive
similar answers \citep{typesafe2026jev}. Open-weight decision models such as Clef and \kev{}
followed, trained from an open-weight LLM with proper scoring rules \citep{cloudflare2026clef,palmer2026kev}. OpenAI now also offers \lunad{}, a decision model built on its LLM \luna{}, through its Decisions API
\citep{openai2026decisions}. Decision models thus avoid
generation and decoding, since every decision comes with a probability in a single request. Whether they are also
robust to how the table and the question are written is open, since a decision model also reads the table as text.

Of the 112 arXiv papers on decision models published in the three weeks after the release of \jev{}, none studies them as a method for tabular data (Table~\ref{tab:census}). For LLMs on tables, decisions are known to depend on the
serialization \citep{sui2024tablemeetsllm,hegselmann2023tabllm}, on the order of rows and columns
\citep{liu2024rethinking,zhou2024freb} and on the in-context examples \citep{hegselmann2023tabllm}. These questions
have not been asked of decision models.

To study them in depth, we change how each request is written without changing the information the decision needs, for example by renaming the options or writing the table in another format.
We call such a change a \emph{rewrite}.
We send every request in its original form and in each rewritten form to decision models and LLMs.

Our research questions are as follows:
\begin{enumerate}[label=\textbf{RQ\arabic*},leftmargin=*,noitemsep,topsep=2pt] \item Are decision models practical for
tabular decisions? \item Are they robust to the formulation of the question, such as the names and order of the
options? \item Are they robust to how the table is organized, such as its format and the order of its rows and
columns? \item Does their confidence indicate which decisions are sensitive to rewrites? \end{enumerate} We evaluate seven models---four decision models (\jev{}, Clef, \kev{} and \lunad{}) and three LLMs (\luna{}, \qwen{}
and \llama{})---across three tabular tasks covering row classification, fact verification and anomaly detection, under 21 types of rewrite. Clef and \kev{} are trained from \qwen{}, which lets us compare
the two ways of deciding on the same backbone (Section~\ref{sec:study}).

Our main findings are as follows:

Decision models perform as well as LLMs on these tasks and are cheaper to run
(Table~\ref{tab:glance}). They are also far less affected by how the question is formulated. In contrast, \lunad{} is as sensitive to the
order of the options as the LLMs. Writing the table differently, however, changes many decisions of every model, even
though accuracy stays almost the same. The decisions that a decision model gives low probability are the ones that
change most often.

\section{Experimental setup}\label{sec:study}

\subsection{Decision interfaces}\label{sec:interfaces}

We refer to a model's output for one item as a \emph{decision}: the class of a sample, the verdict on a
claim about a table, or whether a sample is anomalous. A \emph{request} is one call to a model and may contain several
items. A decision model receives a request as a \emph{state}, such as a group of samples or a table, with one question
per item whose possible answers are fixed in advance. It returns a probability for every option of every question in a single request \citep{typesafe2026jev}. An
LLM receives a prompt built from the same request and generates text, which we map to one of the options. Its
confidence has to be read from that text, from its token probabilities or from repeated draws
\citep{li2026llmasjev,deusser2026jev}.

\subsection{Tasks and models}\label{sec:tasks}

\begin{table}[t]
\centering\small
\setlength{\tabcolsep}{3pt}
\begin{tabularx}{\columnwidth}{@{}>{\raggedright\arraybackslash}p{1.45cm}>{\raggedright\arraybackslash}X>{\raggedright\arraybackslash}p{1.5cm}>{\raggedright\arraybackslash}p{1.75cm}@{}}
\toprule
Task & Input & Decision & Data \\
\midrule
Row classification & One sample and a description of its features & Its class & 10 datasets, 3,481 rows \\
\addlinespace[2pt]
Fact verification & A table and a claim about it & Supported or refuted & TabFact, 2,000 claims \\
\addlinespace[2pt]
Anomaly detection & One sample, a description of the data and statistics of normal samples & Normal or anomalous & ReTabAD, 20 sources, 5,059 samples \\
\bottomrule
\end{tabularx}
\caption{The three tasks: row classification on the ten datasets of TabLLM and FeatLLM
\citep{hegselmann2023tabllm,han2024featllm}, fact verification on TabFact \citep{chen2020tabfact} and anomaly
detection on ReTabAD \citep{retabad2025}. Metrics: Macro-F1, accuracy and AUROC, averaged over datasets or sources.}
\label{tab:tasks}
\end{table}

\begin{figure}[t]
\centering\footnotesize
\renewcommand{\arraystretch}{1.2}
\setlength{\tabcolsep}{3pt}
\begin{tabularx}{\columnwidth}{@{}l>{\raggedright\arraybackslash\ttfamily}X@{}}
\toprule
\multicolumn{2}{@{}l}{\textit{Question}, options with their descriptions} \\
original & \{no, yes\} with descriptions \\
options reversed & \{yes, no\} \\
renamed A, B & \{A, B\}, descriptions kept \\
synonyms & \{false, true\} \\
yes/no question & "Is the target class yes?" \\
paraphrased & "Determine the class \dots" \\
\midrule
\multicolumn{2}{@{}l}{\textit{Row}, columns Recency, Frequency, Monetary, Time} \\
array (original) & [21.0, 3.0, 750.0, 70.0] \\
columns shuffled & [70.0, 21.0, 3.0, 750.0] \\
columns reversed & [70.0, 750.0, 3.0, 21.0] \\
list & - Recency: 21.0 \dots \\
sentences & The Recency is 21.0. \dots \\
JSON object & \{"Recency":21.0, \dots\} \\
Markdown table & | Recency | Frequency | \dots{} | \\
\bottomrule
\end{tabularx}
\caption{One row-classification request (blood donation) under the rewrites of Figure~\ref{fig:edits}, shortened to
the part that each rewrite changes. Shuffling or reversing the columns permutes their names, descriptions and values
together.}
\label{fig:example}
\end{figure}

\paragraph{Tasks.} We use three tasks (Table~\ref{tab:tasks}). Row classification assigns a class to a sample from the
ten datasets of TabLLM and FeatLLM. Fact verification decides whether a table from TabFact supports a claim. Anomaly
detection decides whether a sample from one of 20 ReTabAD sources is anomalous.

\paragraph{Models.} The four decision models are \jev{} 1.13 \citep{typesafe2026jev}, the open-weight Clef
\citep{cloudflare2026clef} and \kev{} \citep{palmer2026kev}, and \lunad{} \citep{openai2026decisions}. \jev{} is
available only through TypeSafe's API. Clef and \kev{} are both trained from \qwen{} \citep{qwen38}. \lunad{} is built on the LLM \luna{} and served through OpenAI's Decisions API. It accepts requests in the same format as \jev{}. OpenAI does
not document how the API computes its probabilities, and we use its public-beta snapshot of 6 October 2026. The LLMs
are \luna{} \citep{openai2026luna} and the open-weight \qwen{} and \llama{} \citep{touvron2024llama3}. We run \luna{} at temperature 1 and decode the open-weight LLMs greedily. Appendix~\ref{app:models} lists all versions.

In row classification, a decision-model request contains the 16 rows of a group as separate typed questions over a
shared state, whereas an LLM prompt contains one row, as in TabLLM. In fact verification and anomaly detection, every
model receives the same claim or the same batch of 15 samples. All decision models receive identical requests, and the LLM prompts are built from them.

\begin{figure*}[t]
\centering
\includegraphics[width=\textwidth]{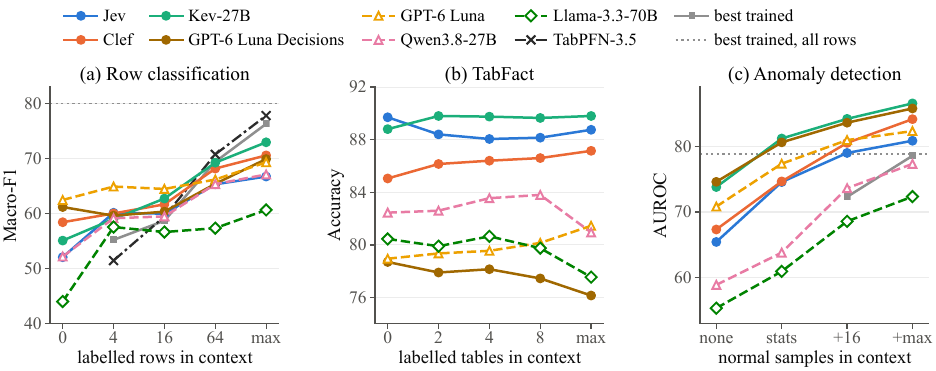}
\caption{In-context examples (RQ1). Grey lines show the best of eight classifiers or five anomaly detectors
trained on the same rows, and dotted lines the best trained on all training rows. \emph{max} stands for as many rows as fit in the context. In (c), \emph{stats} is the original request of ReTabAD, which summarizes the normal training samples by
statistics. \emph{none} leaves these statistics out, and \emph{+16} and \emph{+max} add raw normal samples.}
\label{fig:r3}
\end{figure*}

\subsection{Evaluation}\label{sec:measures}\label{sec:rewrites}

We evaluate row classification by Macro-F1, fact verification by accuracy and anomaly detection by AUROC, each
averaged over datasets or sources (Table~\ref{tab:tasks}). We also report the latency and price of each decision, and
for RQ1 we vary the number of in-context examples from none to as many as fit.

Besides its original form, we send every request under rewrites, which change how it is written but keep all the information the
decision needs (Figure~\ref{fig:example}). Rewrites of the question serve RQ2, rewrites of the table serve RQ3, and
each task uses those of the 21 types of rewrite that apply to it (Table~\ref{tab:rewrites}). We checked that each
rewrite keeps every cell and header by recovering the original request from it. The one
exception is the order of rows on TabFact, which can change the truth of claims about positions such as ``first''. We
leave these claims out of RQ3 and RQ4.

A decision \emph{flips} when it differs from the decision under the original request, and an invalid output counts as
a flip. Even an identical request can receive a different decision. We therefore report each model's \emph{re-send rate}, the share of its decisions that flip when the identical request is sent again, and compare the flips under
rewrites with it. Section~\ref{sec:rq4} defines confidence, and Appendix~\ref{app:results} gives every result with
intervals.

\section{RQ1: Are decision models practical for tabular decisions?}\label{sec:rq1}

\begin{figure*}[t]
\centering
\includegraphics[width=\textwidth]{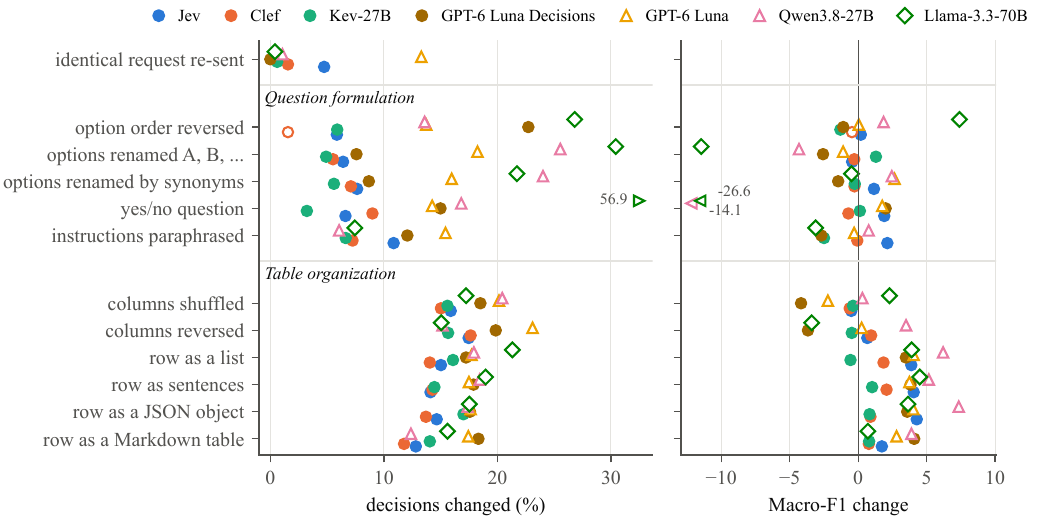}
\caption{Row classification, per rewrite: decisions that differ from the model's original decision (left) and the
change of Macro-F1 (right). The upper block shows RQ2 and the lower block RQ3. Filled markers are decision models and hollow markers LLMs. Clef's point for option order is hollow because Clef sorts the options.}
\label{fig:edits}
\end{figure*}

\begin{table}[t]
\centering\footnotesize
\setlength{\tabcolsep}{3.2pt}
\begin{tabular}{@{}lrrrrrr@{}}
\toprule
& \multicolumn{3}{c}{\textit{performance}} & \multicolumn{3}{c}{\textit{generated tokens}} \\
\cmidrule(lr){2-4}\cmidrule(l){5-7}
Model & Cl. & TF & AD & Cl. & TF & AD \\
\midrule
\jev{} & 52.1 & 89.7 & 74.6 & \multicolumn{3}{c}{\textit{none: one pass}$^b$} \\
Clef & 58.4 & 85.0 & 74.7 & \multicolumn{3}{c}{\textit{none: one pass}} \\
Kev-27B & 55.2 & 88.8 & 81.2 & \multicolumn{3}{c}{\textit{none: one pass}} \\
\lunad{} & 61.2 & 78.7 & 80.5 & \multicolumn{3}{c}{\textit{none}$^c$} \\
\addlinespace[2pt]
\luna{} & 62.5 & 79.0 & 77.3 & 5.6 & 11.8 & 87.5 \\
Qwen3.8-27B & 52.2 & 82.5 & 71.0 & 2.6 & 9.1 & 111.7 \\
Llama-3.3-70B & 44.0 & 80.5 & 61.0 & 2.5 & 8.0 & 44.4 \\
\midrule
\textit{trained}$^a$ & 80.0 & n/a & 78.8 & \multicolumn{3}{c}{\textit{none}} \\
\bottomrule
\end{tabular}
\caption{RQ1: performance under the original request and the tokens each model generates per decision (an anomaly batch carries 15 decisions). Cl. = row classification (Macro-F1), TF = TabFact (accuracy), AD = anomaly detection (AUROC). $^a$Best of eight classifiers trained on the whole training split, LightGBM (Cl.), and of five anomaly detectors, $k$-nearest neighbours (AD, 19 of 20 sources). We train no model for TabFact. $^b$As stated by its provider. \jev{}'s architecture is not public. $^c$No output tokens are billed.}
\label{tab:rq1}
\end{table}

\paragraph{Performance.} Decision models perform as well as LLMs and better on two of three tasks. \jev{} verifies
TabFact claims best and \kev{} detects anomalies best, while \luna{} leads on row classification
(Table~\ref{tab:rq1}). Clef and \kev{} are also more accurate than \qwen{}, which shares their backbone, on every task. \lunad{} comes within 1.3 points of \luna{} on row classification and TabFact and is 3.2 points
above it on anomaly detection.
In-context examples help the decision models as they help the LLMs. With as many rows as fit in the context,
Clef and \kev{} beat every LLM on row classification and every LLM and detector fitted to the same samples on anomaly
detection (Figure~\ref{fig:r3}). From about 64 rows on, TabPFN-3.5 \citep{priorlabs2026tabpfn35} and methods trained
on the same rows match or beat every in-context model on row classification.

\paragraph{Latency and price.} The LLMs generate 2.5 to 112 tokens per decision, and the other models none
(Table~\ref{tab:rq1}). Of the closed models, \jev{} and \lunad{} take 0.08--0.11 seconds per row or sample, against 0.73--1.62 for
\luna{} (Table~\ref{tab:cost}). Per decision, \luna{} costs 1.6 to 8 times as much as \jev{}. Served through the
Decisions API, the same model costs 3.4 times less on row classification and 1.8 times less on anomaly detection. On
TabFact, where every request holds a single claim, the two interfaces cost the same per decision. We time only the closed models, since the open-weight models run on our own servers.

\takeaway{Decision models are practical for tabular decisions. They perform as well as LLMs or better, and are generally faster and cheaper per decision. Where 64 or more labelled rows exist, a tabular foundation model or a trained model is better.}

\section{RQ2: Are decision models robust to the formulation of the question?}\label{sec:rq2}

\paragraph{Decision models are robust to the names and order of the options.} Renaming or reordering the options
changes 2--8\% of the classifications of \jev{}, Clef and \kev{}, at most 5.5 percentage points above their re-send rates
(Figure~\ref{fig:edits}, upper block). Their Macro-F1 changes by at most 1.3 points. The same rewrites change 14--30\%
of the open-weight LLMs' classifications and cost them up to 11.5 points. Posing the question as yes/no costs these three models at most 2 points but \qwen{} 14, and \llama{} answers it with ``Yes.'' so often that half of its outputs are
unusable. Only paraphrasing the instructions affects both kinds of model alike. On TabFact these rewrites change under 5\% of their verdicts and under 7\% of the open-weight LLMs'. Clef and \kev{} change far fewer decisions under
these rewrites than \qwen{}, which shares their backbone. For option order, part of this comes from Clef's release
code, which sorts the options.

\begin{table}[t]
\centering\footnotesize
\setlength{\tabcolsep}{3.2pt}
\begin{tabular}{@{}lrrrrrr@{}}
\toprule
& \multicolumn{3}{c}{\textit{seconds}} & \multicolumn{3}{c}{\textit{US\$ / million}} \\
\cmidrule(lr){2-4}\cmidrule(l){5-7}
Model & Cl. & TF & AD & Cl. & TF & AD \\
\midrule
\jev{} & 0.08 & 1.33 & 0.09 & 10 & 38 & 16 \\
\lunad{} & 0.10 & 1.60 & 0.11 & 24 & 61 & 37 \\
\luna{} & 1.62 & 1.59 & 0.73 & 81 & 62 & 68 \\
\bottomrule
\end{tabular}
\caption{Latency and price per decision of the closed models: mean wall-clock seconds, and US\$ per million decisions as billed, on the original requests. A \jev{} or \lunad{} request answers 16 rows or a batch of 15 samples, and a \luna{} request one row, one claim or one batch.}
\label{tab:cost}
\end{table}

\paragraph{\lunad{} is not robust to the order of the options.} \lunad{} repeats its decision whenever the
identical request is re-sent. Yet reversing the options changes 23\% of its classifications, 12\% of its TabFact
verdicts and 22\% of its anomaly decisions, and renaming them changes 8--17\%. Few of these flips come from ties of
its rounded probabilities. Even among its classifications with a probability of at least 0.8, 14\% flip, and 92--100\%
of all flips move to the option now listed first. Robustness to the question thus does not come with the decision
interface itself.

\takeaway{Decisions of \jev{}, Clef and \kev{} remain largely unchanged when the options are renamed or reordered. However, LLMs and \lunad{} can change up to 30\% of their decisions.}

\section{RQ3: Are decision models robust to how the table is organized?}\label{sec:rq3}

\begin{table}[t]
\centering\footnotesize
\setlength{\tabcolsep}{2.4pt}
\begin{tabular}{@{}lrrr@{}}
\toprule
Model & Row classif. & TabFact & Anomaly det. \\
\midrule
\jev{} & 33.8 & 19.3 & 21.1 \\
Clef & 31.7 & 23.0 & 25.8 \\
Kev-27B & 32.2 & 18.3 & 17.7 \\
\lunad{} & 39.6 & 38.3 & 29.9 \\
\addlinespace[2pt]
\luna{} & 42.8 & 60.8 & 63.4 \\
Qwen3.8-27B & 31.8 & 27.4 & 23.3 \\
Llama-3.3-70B & 35.5 & 33.7 & 53.4 \\
\bottomrule
\end{tabular}
\caption{Items (\%) whose decision changes under at least one information-preserving rewrite of the table (6, 10 and 8 types of rewrite, Table~\ref{tab:rewrites}). \luna{} is run at temperature 1, so its shares include its re-send rate.}
\label{tab:changed}
\end{table}

\paragraph{No language model is fully robust to how the table is organized.} Writing each row as a list, as sentences,
as a JSON object or as a Markdown table instead of an array of values changes 12--21\% of the row classifications of
every model. This is about ten times the re-send rate of the open-weight models. Shuffling or reversing the columns changes
15--23\% (Figure~\ref{fig:edits}, lower block). On TabFact, other table formats and row or column orders change 3--13\% of the decision models' verdicts and 4--12\% of the open-weight LLMs'. Here decision models
and LLMs do not differ. Clef and \kev{} change about as many decisions as \qwen{}, which shares their backbone.

\paragraph{Nor is any model fully robust to how the in-context examples are presented.} We present the same 16 in-context examples with their classes interleaved instead of grouped, or as text lines instead of arrays. This changes 7--12\% of
the decision models' classifications, 7--8\% of \qwen{}'s and 15\% of \llama{}'s, far above their re-send rates.
Macro-F1 changes by at most 3 points. \luna{}'s 14\% is close to its re-send rate. The presentation of the in-context examples is thus part of the organization of the table as well.

\begin{table}[t]
\centering\small
\setlength{\tabcolsep}{4pt}
\begin{tabular}{@{}llrrr@{}}
\toprule
Model & Confidence & Cl. & TF & AD \\
\midrule
\multicolumn{5}{@{}l}{\textit{Trained from the same backbone}} \\
\quad Clef & probability & 1.4 & 4.1 & 0.7 \\
\quad Kev-27B & probability & 3.1 & 2.5 & 0.1 \\
\quad Qwen3.8-27B & ten draws & 2.6 & 9.7 & 4.8 \\
 & stated & 23.4 & 29.2 & 9.2 \\
\midrule
\multicolumn{5}{@{}l}{\textit{The same model through two interfaces}} \\
\quad \lunad{} & probability & 1.8 & 15.5 & 4.2 \\
\quad \luna{} & ten draws & 20.2 & 48.7 & 47.6 \\
 & stated & 22.5 & 61.6 & 50.0 \\
\midrule
\multicolumn{5}{@{}l}{\textit{Other models}} \\
\quad \jev{} & probability & 0.9 & 4.6 & 0.6 \\
\quad Llama-3.3-70B & ten draws & 26.8 & 27.5 & 45.9 \\
 & stated & 26.8 & 27.7 & 39.6 \\
\bottomrule
\end{tabular}
\caption{Of each model's 25\% most confident items, the share (\%) changed by at least one rewrite of the table. Cl. = row classification, TF = TabFact, AD = anomaly detection.}
\label{tab:rq4}
\end{table}

\paragraph{Individual decisions change while overall accuracy stays the same.}
For every model, some rewrite of the table changes the decision on about a third of the row classifications and on a
fifth or more of the TabFact claims and anomaly-detection samples (Table~\ref{tab:changed}). Re-sending the request changes only 2--9\% of the open-weight models' rows. Yet Macro-F1 changes by less than a point for some models and by up
to 7 for others, because the flips fall on borderline items that are right about as often as wrong
(Section~\ref{sec:rq4}). This is the arbitrariness that the multiplicity literature studies across retrained models
and prompts \citep{marx2020predictive,ganesh2025prompt}, here caused by the table.

\takeaway{Neither decision models nor LLMs are fully robust to how the table is organized. About a third of every model’s row-classification decisions change under at least one rewrite of the table. But accuracy largely hides these changes.}

\section{RQ4: Does their confidence indicate which decisions are sensitive to rewrites?}\label{sec:rq4}

Every model we test makes some decisions that a rewrite of the table changes (Section~\ref{sec:rq3}). We call such a decision \emph{fragile}, meaning that at least one rewrite of the table flips it, and any other decision \emph{stable}. We ask whether a model's confidence indicates which of its decisions are fragile. Since a table can
hold many rows, such a signal is most useful if it comes with each decision rather than from repeated requests.

\paragraph{Confidence.}\label{sec:confidence} Let $d$ be a model's original decision on an item. A decision model returns a probability $p(o)$ for every option $o$, and its confidence is the probability of the
chosen option,
\begin{equation}
  c = \max_o p(o).
\end{equation}
This is not the \texttt{confidence} field that some of these APIs return, which is the gap between the two highest
probabilities. The closed LLM returns no probabilities, so for all three LLMs we draw ten answers $a_1, \dots, a_{10}$
at temperature 1 and use the share that agrees with $d$,
\begin{equation}
  c = \tfrac{1}{10}\,\bigl|\{k : a_k = d\}\bigr|,
\end{equation}
where an invalid draw counts as different. We also compare two other LLM readouts, the stated confidence divided by
100 and the probability of the first token of the generated answer (Table~\ref{tab:readout}).

\paragraph{Measure.} The \emph{flip AUROC} measures how well low confidence ranks the decisions that a rewrite
flips. It is the AUROC of $-c$ for a flip over all pairs of an item and a rewrite. For comparison, the \emph{error AUROC} measures how well $-c$ ranks wrong decisions. For both, 0.5 is chance.

\paragraph{A decision model's confidence is a strong indicator of fragile decisions.} Of the 25\% of items with its highest confidence, 0.1--5\% change under some rewrite of the table, against
about a third of all items (Table~\ref{tab:rq4}). The exception is \lunad{} on TabFact, with 15.5\%. Pooled over the rewrites both kinds of model receive, the flip AUROC is 0.82--0.87 for the decision models on row
classification, 0.81--0.90 on TabFact and 0.84--0.93 on anomaly detection, with \lunad{} lowest on every task. It is similar for each
rewrite on its own (Figure~\ref{fig:byedit}). Their mean confidence on a dataset also ranks the ten classification datasets by flip rate (Spearman $-$0.62 to $-$0.90, Figure~\ref{fig:datasets}), as does an LLM's ten-draw
agreement.

\paragraph{An LLM's confidence is a weaker indicator of fragile decisions.} Read as the share of ten draws that repeat
the original decision, it ranks the flips less well than the probability of \jev{}, Clef and \kev{} on every task, and the best LLM
ranks them about as well as \lunad{} (Table~\ref{tab:auroc}). The
LLMs' most confident decisions are also less stable (Table~\ref{tab:rq4}). The confidence an LLM states is the least
informative of all. With the best of its three readouts, the best LLM comes within 0.02 of the weakest of \jev{}, Clef and \kev{} on row classification and TabFact and stays 0.05 behind on anomaly detection (Table~\ref{tab:readout}), but these
readouts need ten decoding passes or token probabilities.

\begin{figure}[t]
\centering
\includegraphics[width=\columnwidth]{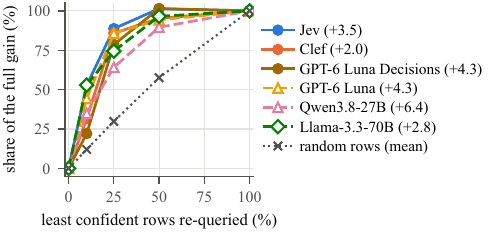}
\caption{Re-querying the least confident rows (row classification). For the x\% of rows with the lowest confidence,
the answer in the original format is replaced by the combination of the answers in the original and the
four other formats. The y-axis gives the share of the Macro-F1 gain obtained when every row is re-queried, and the
legend gives that gain in points.}
\label{fig:selective}
\end{figure}

\paragraph{Re-querying the least confident rows.} Since low confidence marks the decisions that a rewrite is likely to
change, it can also tell which rows are worth asking again. On row classification, we send a share of the rows again
in the four other formats of Table~\ref{tab:rewrites} and combine the five answers, by the highest mean probability
for a decision model and by majority vote for an LLM. Re-querying every row raises Macro-F1 by 2.0 to 6.4 points for
six of the models but takes four more requests per row. Re-querying only the 25\% least confident rows takes one more
request per row on average and recovers 79--89\% of the gain for the decision models and 64--86\% for the LLMs,
against 23--44\% for random rows (Figure~\ref{fig:selective}). Most of this gain comes from leaving the original array
of values, the weakest format for six of the seven models, and the best other format alone comes within 0.2 points of
the combination or exceeds it. The rows whose decisions change with the format are thus mostly the least confident
ones, and re-querying them undoes most of the effect of a weak format. \kev{}, whose weakest format is not the array,
loses 0.7 points (not shown).

\paragraph{Confidence follows stability more than correctness.} Confident decisions are more often right. Among the
25\% of decisions with the highest confidence, the decision models are right 67--99\% of the time, against 43--73\%
among the 25\% with the lowest. Yet their confidence separates stable from fragile decisions far better than right
from wrong ones (Table~\ref{tab:confcorrect}). On row classification and anomaly detection, stable but wrong decisions
carry almost as much confidence as stable and right ones (0.67--0.86 against 0.76--0.90) and far more than fragile
ones (0.54--0.66). On TabFact the order is the same, with smaller gaps. Such stable errors make up 17--23\% of the
decisions in row classification and anomaly detection and 4--5\% on TabFact, and no signal of instability can reveal
them \citep[cf.][]{ganesh2026rethinking}. \llama{}'s confidence is high in all four groups (0.85--1.00).

\paragraph{Why confidence predicts flips.} A decision flips when a rewrite moves its \emph{margin}, the gap between the
log-probabilities of the chosen option and the best other option, below zero
\citep[cf.][]{cohen2019smoothing,bhojanapalli2021reproducibility}. For the decision models, the flip rates predicted
from the margin alone, assuming a small shift that is nearly independent of the item, match the observed ones decile
by decile (Figure~\ref{fig:noise}). Such a shift flips only borderline decisions, which would explain why the flips
cancel in accuracy and why the confidence predicts them.

\takeaway{A decision model's probability indicates which of its decisions a rewrite is likely to change. The confidence of an LLM is a weaker indicator. Re-querying only the least confident rows recovers most of the gain of re-querying every row.}

\section{Related work}\label{sec:related}

\paragraph{LLMs on tables.} LLMs can classify rows serialized as text from a few labelled examples, by prompting or
fine-tuning \citep{dinh2022lift,hegselmann2023tabllm}. They can also build features and rules for a conventional model
\citep{han2024featllm,ye2025delta}. LLMs also detect anomalous samples
\citep{tsai2025anollm,retabad2025,ye2026tadsurvey} and verify claims about tables on TabFact \citep{chen2020tabfact}, and
\citet{fang2024llmtabular} survey the field. Each of these methods addresses one kind of task and reports accuracy. We
compare LLMs with decision models on three kinds and look at single decisions.

\paragraph{Robustness of LLMs.} LLM answers change with the order, names and format of the options and with the
wording of the prompt \citep{zhao2021calibrate,zheng2024mcq,sclar2024quantifying,mizrahi2024multiprompt}. On tables,
the serialization changes results \citep{sui2024tablemeetsllm,fang2024llmtabular}, and reordering or transposing a
table lowers accuracy on table question answering \citep{liu2024rethinking,zhao2023robut,zhou2024freb}. For single
predictions, \citet{hamman2025tabular} find that the probability of tabular LLMs does not reveal their inconsistency
across fine-tuning seeds, and \citet{ganesh2026rethinking} find that an LLM's uncertainty identifies prompt-sensitive
items but not wrong ones. We measure which individual decisions rewrites of the table change, for LLMs and for models
that return a probability by design.

\paragraph{Decision models.} Early studies measure \jev{}'s accuracy and calibration on text \citep{deusser2026jev},
its consistency under benign and injected edits \citep{hu2026jevadvbench} and the readout stability of open-weight
decision models \citep{li2026readout}. Others apply decision models to single tabular applications
\citep{severo2026jevids,jung2026hvac}, and an audit lists the missing tests \citep{tang2026audit}. We compare decision
models, including one built on an LLM, with LLMs across tabular tasks.

\section{Discussion}\label{sec:agenda}

\paragraph{When to use decision models?} Decision models suit tabular decisions whose answers form a fixed set, such
as classifying rows, verifying claims or flagging anomalies. In tasks with few labelled rows, they perform as well as LLMs, cost
less per decision, respond faster and return a useful confidence with each decision (Section~\ref{sec:rq1}). They are most useful when many
rows need a decision, when the wording of the question may vary and when uncertain decisions need to be identified
(Sections~\ref{sec:rq2} and~\ref{sec:rq4}). They are less suited when many labelled rows are available. From around 64 labelled rows on, tabular foundation models and trained models perform better (Section~\ref{sec:rq1}), and decision models have shorter context windows than LLMs, which limits how many labelled rows they can read. By design they return only
one of a fixed set of choices, so they do not cover tasks that need free text, such as an explanation or a computed
value. 

\paragraph{When to use LLMs?} LLMs suit the task better when the answer has to be written in tokens rather than chosen from a fixed set, for example when a decision must be explained or an open question about a table answered. A strong LLM is also a reasonable starting point when almost no labelled rows exist (Section~\ref{sec:rq1}). When an LLM is used for decisions with a fixed set of answers, a single question prompt may not be relied on, since presenting the options in several orders and aggregating the answers can reduce its dependence \citep{zheng2024mcq}.

\paragraph{Future directions.} No model we test is robust to how the table is organized, and this is the main
limitation of decision models on tables. A natural next step is to train decision models on rewrites of the table. In previous work, LLMs have been trained to predict on unseen tables in few shots more accurately than XGBoost and TabPFN \citep{gardner2024tabula}. With RLCD, decision 
models can also follow this path and be trained  to improve their decision accuracy and robustness on tabular tasks. 
LLMs and decision models can also work together: a decision model can make most of the decisions, and an LLM can explain or review the few it is least confident about. Systems that combine the two kinds of model in this way are another direction for future work.

\section{Conclusion}\label{sec:conclusion}

We studied typed decision models as a method for decisions on tabular data. On row classification, fact verification
and anomaly detection, they perform as well as LLMs at a fraction of the cost. Except for \lunad{}, renaming or reordering the options barely changes their decisions. How the table is organized, however, affects decision models and LLMs alike, and about
a third of the row decisions change under some rewrite of the table. The decisions to which a decision model gives low
probability are the ones most likely to change. Whether decision models can be made robust to how the table is organized remains open.

\section*{Limitations}

First, each of the seven models is tested in one version. \jev{} and \lunad{} are closed services whose
internal processing is not visible to us and may change across versions, and the Decisions API is still in public
beta. The two open-weight decision models reproduce the findings we rely on, but other decision models, and later versions of
the ones we test, may behave differently.

Second, we cover three kinds of tabular decisions and use subsets of each benchmark, since every request is sent under
every rewrite to seven models. The 21 types of rewrite cover the format of the table, the order of its rows and
columns and the formulation of the question. They do not cover other changes that keep the content, such as renaming
columns or converting units, so our results say little about robustness to those changes.

\bibliography{references}

\appendix

\twocolumn[{%
\section{Models}\label{app:models}

Table~\ref{tab:models} lists the models we evaluate, the exact versions and how we run them.

\vspace{6pt}
{\centering\footnotesize
\renewcommand{\arraystretch}{1.1}
\begin{tabularx}{\textwidth}{@{}l>{\raggedright\arraybackslash}p{2.9cm}>{\raggedright\arraybackslash}X
>{\raggedright\arraybackslash}p{3.6cm}@{}}
\toprule
Model & Access & Version & How we run it \\
\midrule
\multicolumn{4}{@{}l}{\textit{Decision models, whose confidence is the probability they return}} \\
\addlinespace[1pt]
\jev{} & Closed & \texttt{typesafe/jev-1.13}\hfil\break revision \texttt{20260917} & TypeSafe's API through OpenRouter \\
Clef & Open-weight, trained from \qwen{} & \texttt{Cloudflare/clef}\hfil\break commit \texttt{2f3de3dd} & Our implementation of its release code \\
\kev{} & Open-weight, trained from \qwen{} & \texttt{jaredpalmer/kev-27b}\hfil\break revision \texttt{28be62e9} & Official server, bf16, 4$\times$A800 \\
\lunad{} & Closed, built on\hfil\break \luna{} & \texttt{openai/gpt-6-luna-decisions}\hfil\break snapshot \texttt{gpt-6-luna-decisions-20261006} & OpenAI's Decisions API (public beta) through OpenRouter \\
\midrule
\multicolumn{4}{@{}l}{\textit{LLMs, whose confidence is the share of ten draws that repeat the original decision}} \\
\addlinespace[1pt]
\luna{} & Closed & \texttt{openai/gpt-6-luna}\hfil\break snapshot \texttt{gpt-6-luna-20260922} & OpenAI's API through OpenRouter, temperature 1, reasoning off \\
\qwen{} & Open-weight & \texttt{Qwen/Qwen3.8-27B}\hfil\break commit \texttt{1d4bf0f2} & vLLM, greedy decoding \\
\llama{} & Open-weight & \texttt{meta-llama/Llama-3.3-70B-Instruct}\hfil\break weights checked by sha256 against the official release & vLLM, greedy decoding \\
\bottomrule
\end{tabularx}
\captionof{table}{The models we evaluate. Section~\ref{sec:rq4} defines the confidence. For every LLM, the ten draws are
taken at temperature 1.}\label{tab:models}
\par}
\vspace{14pt}
}]

\section{Compute}\label{app:compute}

Table~\ref{tab:compute} lists the hardware and the dates of all runs. The open-weight models run on servers with four
or six NVIDIA A800 80GB PCIe GPUs and on a server with four NVIDIA RTX 6000D GPUs (84 GB each). The trained methods
and TabPFN-3.5 run on CPUs.

\begin{center}
\footnotesize
\setlength{\tabcolsep}{4pt}
\begin{tabular}{@{}lll@{}}
\toprule
Model & Hardware & Dates \\
\midrule
\jev{} & API & 1--5 Oct \\
Clef & A800, RTX 6000D & 1--9 Oct \\
\kev{} & A800 & 6--9 Oct \\
\lunad{} & API & 12 Oct \\
\luna{} & API & 1--10 Oct \\
\qwen{} & A800, RTX 6000D & 1--9 Oct \\
\llama{} & A800 & 7--9 Oct \\
Trained methods & CPU & 5--9 Oct \\
TabPFN-3.5 & CPU & 9 Oct \\
\bottomrule
\end{tabular}
\captionof{table}{Hardware and dates of the runs (2026, UTC). The closed models are called through their providers' APIs.}
\label{tab:compute}
\end{center}

\newpage
\section{Further results}\label{app:results}

\makeatletter
\setlength{\@fptop}{0pt}\setlength{\@dblfptop}{0pt}
\makeatother
\renewcommand{\dbltopfraction}{0.95}
\renewcommand{\dblfloatpagefraction}{0.5}
\setcounter{dbltopnumber}{3}

\begin{figure}[!ht]
\centering
\includegraphics[width=0.92\columnwidth]{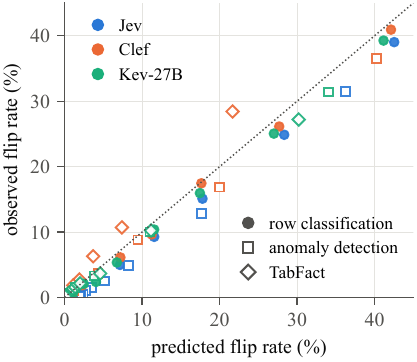}
\caption{Decision models: observed flip rate against the rate predicted when a rewrite adds item-independent noise to
the margin, per margin decile and task.}
\label{fig:noise}
\end{figure}

\begin{table*}[t]
\centering\footnotesize
\setlength{\tabcolsep}{3.6pt}
\begin{tabular}{@{}lllllll@{}}
\toprule
& \multicolumn{3}{c}{\textit{predicts a flip}} & \multicolumn{3}{c}{\textit{predicts an error}} \\
\cmidrule(lr){2-4}\cmidrule(l){5-7}
Model & Row classif. & Anomaly & TabFact & Row classif. & Anomaly & TabFact \\
\midrule
\jev{} & 0.87 {\scriptsize [0.86, 0.88]} & 0.89 {\scriptsize [0.89, 0.90]} & 0.89 {\scriptsize [0.88, 0.91]} & 0.59 & 0.65 & 0.82 \\
Clef & 0.85 {\scriptsize [0.85, 0.86]} & 0.89 {\scriptsize [0.88, 0.89]} & 0.87 {\scriptsize [0.86, 0.89]} & 0.67 & 0.61 & 0.84 \\
Kev-27B & 0.85 {\scriptsize [0.84, 0.86]} & 0.92 {\scriptsize [0.92, 0.93]} & 0.90 {\scriptsize [0.88, 0.91]} & 0.70 & 0.70 & 0.84 \\
\shortstack[l]{GPT-6 Luna\\Decisions} & 0.82 {\scriptsize [0.81, 0.83]} & 0.83 {\scriptsize [0.83, 0.84]} & 0.81 {\scriptsize [0.80, 0.83]} & 0.74 & 0.70 & 0.78 \\
\midrule
\luna{} & 0.83 {\scriptsize [0.82, 0.84]} & 0.71 {\scriptsize [0.70, 0.72]} & 0.73 {\scriptsize [0.72, 0.75]} & 0.68 & 0.62 & 0.69 \\
Qwen3.8-27B & 0.77 {\scriptsize [0.76, 0.78]} & 0.84 {\scriptsize [0.83, 0.85]} & 0.83 {\scriptsize [0.82, 0.85]} & 0.70 & 0.62 & 0.78 \\
Llama-3.3-70B & 0.64 {\scriptsize [0.63, 0.65]}$^\dagger$ & 0.67 {\scriptsize [0.66, 0.68]} & 0.66 {\scriptsize [0.64, 0.68]} & 0.55 & 0.51 & 0.57 \\
\bottomrule
\end{tabular}
\caption{How well a model's confidence in its original decision predicts (left) that an information-preserving rewrite changes the decision and (right) that the original decision is wrong (flip AUROC and error AUROC; 0.5 is chance). Left: pooled over the rewrites of Figure~\ref{fig:edits} and the table formats, with 95\% cluster-bootstrap intervals. Confidence: for the decision models, the returned probability of the chosen option; for the LLMs, the share of ten draws at temperature 1 that give the original decision. $^\dagger$Strict parse: an answer such as ``Yes.'' to the yes/no question counts as invalid, hence as changed. Ignoring case and a final full stop: Llama-3.3-70B on row classif. 0.66 [0.65, 0.67].}
\label{tab:auroc}
\end{table*}

\begin{table}[!ht]
\centering\footnotesize
\setlength{\tabcolsep}{2.2pt}
\begin{tabular}{@{}lccc@{}}
\toprule
Model & Ten draws & Stated & Answer prob. \\
\midrule
\multicolumn{4}{@{}l}{\textit{Row classification}} \\
\quad \luna{} & \textbf{0.83} & 0.72 & 0.71 \\
\quad Qwen3.8-27B & \textbf{0.77} & 0.60 & 0.65 \\
\quad Llama-3.3-70B & 0.64 & 0.50 & \textbf{0.73} \\
\multicolumn{4}{@{}l}{\quad\textit{weakest of \jev{}, Clef and \kev{}: Kev-27B, 0.85}} \\
\multicolumn{4}{@{}l}{\quad\textit{\lunad{}: 0.82}} \\
\midrule
\multicolumn{4}{@{}l}{\textit{TabFact}} \\
\quad \luna{} & \textbf{0.73} & 0.58 & -- \\
\quad Qwen3.8-27B & 0.83 & 0.52 & \textbf{0.86} \\
\quad Llama-3.3-70B & 0.66 & 0.56 & \textbf{0.83} \\
\multicolumn{4}{@{}l}{\quad\textit{weakest of \jev{}, Clef and \kev{}: Clef, 0.87}} \\
\multicolumn{4}{@{}l}{\quad\textit{\lunad{}: 0.81}} \\
\midrule
\multicolumn{4}{@{}l}{\textit{Anomaly detection}} \\
\quad \luna{} & \textbf{0.71} & 0.65 & -- \\
\quad Qwen3.8-27B & \textbf{0.84} & 0.72 & 0.71 \\
\quad Llama-3.3-70B & \textbf{0.67} & 0.59 & 0.61 \\
\multicolumn{4}{@{}l}{\quad\textit{weakest of \jev{}, Clef and \kev{}: Clef, 0.89}} \\
\multicolumn{4}{@{}l}{\quad\textit{\lunad{}: 0.84}} \\
\bottomrule
\end{tabular}
\caption{Every readout of an LLM's confidence, on the same rewrites: flip AUROC (rewrites shared by both kinds of model). \emph{Ten draws}: share of ten draws at temperature 1 that give the original decision, the readout of the main text. \emph{Stated}: the confidence the model gives when asked, 0--100 (anomaly detection: the confidence in its answer). \emph{Answer prob.}: probability of the first token of the generated label, or the stated anomaly score's distance from 0.5; for \luna{}, agreement of three draws. Bold: the best readout of each model and task. --: not available.}
\label{tab:readout}
\end{table}

\begin{table}[!ht]
\centering\footnotesize
\setlength{\tabcolsep}{3pt}
\begin{tabularx}{\columnwidth}{@{}l>{\raggedright\arraybackslash}X@{}}
\toprule
arXiv & Table input (application or setting) \\
\midrule
2609.23136 & telemetry tables (6G network intents) \\
2609.24395 & video feature records (video quality) \\
2609.27535 & survey micro-data as personas (social science) \\
2609.33689 & KPM records (5G network control) \\
2609.36154 & accelerometer features (activity recognition) \\
2609.37470 & joined tables or nested records (synthetic world) \\
2610.00376 & packet features (traffic classification) \\
2610.01079 & NSL-KDD flow records (intrusion detection) \\
2610.02046 & database tuples (semantic queries) \\
2610.03073 & security telemetry (security tasks) \\
2610.08089 & relational tuples (synthetic query plans) \\
2610.08675 & FinQA tables (citation checking) \\
2610.09487 & column names and example values (schema matching) \\
2610.09937 & sensor tables and features (HVAC faults) \\
\bottomrule
\end{tabularx}
\caption{Of the 112 arXiv papers on decision models posted between 15 September and 8 October 2026, the
fourteen whose main input is a table or table rows.}
\label{tab:census}
\end{table}

\begin{table*}[t]
\centering\small
\setlength{\tabcolsep}{5pt}
\begin{tabularx}{\textwidth}{@{}llXl@{}}
\toprule
Task & Code & Rewrite & Source \\
\midrule
Row classification & \texttt{COLS} & Columns shuffled: one fixed permutation per dataset; field descriptions, names and values permuted together & order \\
 & \texttt{COLR} & Columns in reverse order & order \\
 & \texttt{LIST} & One line per field, ``- name: value'' (TabLLM list) & format \\
 & \texttt{TEXT} & One sentence per field, ``The name is value.'' (TabLLM text) & format \\
 & \texttt{JSONOBJ} & JSON object \{name: value\} & format \\
 & \texttt{MD} & One-row Markdown table (header and value row) & format \\
\midrule
TabFact & \texttt{COLS} & Columns shuffled: one fixed permutation per table; header and rows permuted together & order \\
 & \texttt{COLR} & Columns in reverse order & order \\
 & \texttt{ROWS} & Data rows shuffled; header kept first$^b$ & order \\
 & \texttt{ROWR} & Data rows in reverse order$^b$ & order \\
 & \texttt{TRANS} & Table transposed; the format sentence says that each line starts with a column name followed by that column's values & order \\
 & \texttt{MD} & Markdown table & format \\
 & \texttt{CSV} & CSV with standard quoting & format \\
 & \texttt{JSONREC} & List of JSON records, one object per row keyed by the header & format \\
 & \texttt{HTML} & HTML table & format \\
 & \texttt{TAPEX} & TAPEX linearization, ``col : \dots{} row 1 : \dots'' & format \\
\midrule
Anomaly detection & \texttt{COLS} & Fields shuffled within each sample & order \\
 & \texttt{COLR} & Fields in reverse order & order \\
 & \texttt{ROWS} & Samples of a batch shuffled; sample identifiers unchanged & order \\
 & \texttt{ROWR} & Samples of a batch in reverse order & order \\
 & \texttt{JSONOBJ} & Each sample as a JSON object & format \\
 & \texttt{LIST} & Each sample as ``- name: value'' lines & format \\
 & \texttt{TEXT} & Each sample as sentences, ``The name is value.'' & format \\
 & \texttt{MD} & The batch as one Markdown table, sample identifier in the first column & format \\
 & \texttt{DEC3} & Numeric fields rendered with three decimals instead of two$^a$ & format \\
\bottomrule
\end{tabularx}
\caption{Rewrites of the table that keep the information needed for the decision. The original format is, for row classification, field descriptions followed by one array of values; for TabFact, the official format with a header line and \# between cells; for anomaly detection, the serializer of the ReTabAD authors with 15 samples per batch. Source: order of rows and columns, or format of the table.  $^a$Adds a decimal of precision, so it is left out of Tables~\ref{tab:changed} and~\ref{tab:rq4}. $^b$Can change the truth of claims that refer to positions (\emph{first}, \emph{last}, \dots); such claims are left out of Tables~\ref{tab:changed} and~\ref{tab:rq4}.}
\label{tab:rewrites}
\end{table*}

\FloatBarrier

\begin{table*}[t]
\centering\scriptsize
\setlength{\tabcolsep}{2.4pt}
\begin{tabular}{@{}lrrrrrrrrrrrrrr@{}}
\toprule
& \multicolumn{8}{c}{\textit{decision models}} & \multicolumn{6}{c}{\textit{LLMs}} \\
\cmidrule(lr){2-9}\cmidrule(l){10-15}
& \multicolumn{2}{c}{\jev{}} & \multicolumn{2}{c}{Clef} & \multicolumn{2}{c}{Kev-27B} & \multicolumn{2}{c}{\shortstack[l]{GPT-6 Luna\\Decisions}} & \multicolumn{2}{c}{\luna{}} & \multicolumn{2}{c}{Qwen3.8-27B} & \multicolumn{2}{c}{Llama-3.3-70B} \\
\cmidrule(lr){2-3}\cmidrule(lr){4-5}\cmidrule(lr){6-7}\cmidrule(lr){8-9}\cmidrule(lr){10-11}\cmidrule(lr){12-13}\cmidrule(lr){14-15}
Condition & Flip & Score & Flip & Score & Flip & Score & Flip & Score & Flip & Score & Flip & Score & Flip & Score \\
\midrule
\multicolumn{15}{@{}l}{\textit{Row classification (Macro-F1)}} \\
\quad original request & -- & 52.1 & -- & 58.4 & -- & 55.2 & -- & 61.2 & -- & 62.5 & -- & 52.2 & -- & 44.0 \\
\quad identical re-send & 4.7 & 52.4 & 1.6 & 58.0 & 0.6 & 54.5 & 0.0 & 61.2 & 13.3 & 63.0 & 1.1 & 52.9 & 0.4 & 44.1 \\
\quad options in reverse order & 5.9 & 52.3 & 1.6 & 58.0 & 5.9 & 53.9 & 22.7 & 60.1 & 13.7 & 62.5 & 13.6 & 54.1 & 26.8 & 51.4 \\
\quad option names A, B, \dots & 6.4 & 51.7 & 5.5 & 58.2 & 4.9 & 56.5 & 7.6 & 58.6 & 18.2 & 61.4 & 25.5 & 47.9 & 30.4 & 32.6 \\
\quad option names replaced by synonyms & 7.6 & 53.3 & 7.1 & 58.2 & 5.6 & 54.9 & 8.7 & 59.7 & 16.0 & 65.1 & 24.0 & 54.7 & 21.7 & 43.6 \\
\quad no option descriptions & 5.4 & 52.0 & 6.1 & 57.6 & 6.0 & 54.2 & 9.3 & 63.0 & -- & -- & -- & -- & -- & -- \\
\quad original, binary datasets & -- & 54.9 & -- & 62.6 & -- & 59.4 & -- & 64.7 & -- & 65.5 & -- & 59.2 & -- & 46.2 \\
\quad yes/no question & 6.6 & 56.8 & 9.0 & 61.9 & 3.2 & 59.5 & 15.0 & 66.7 & 14.3 & 67.3 & 16.8 & 45.2 & 56.9 & 19.6 \\
\quad paraphrased instructions & 10.9 & 54.2 & 7.2 & 58.4 & 6.6 & 52.7 & 12.1 & 58.5 & 15.4 & 62.2 & 6.1 & 53.0 & 7.4 & 40.9 \\
\cmidrule(l){1-15}
\quad unverified hint appended & 8.8 & 51.3 & 15.6 & 56.6 & 4.7 & 55.1 & 17.6 & 60.1 & 16.2 & 61.1 & 10.5 & 48.0 & 33.9 & 45.9 \\
\quad no task description & 14.6 & 54.2 & 10.7 & 55.6 & 14.1 & 57.4 & 18.6 & 54.2 & 20.1 & 61.4 & 7.2 & 52.5 & 15.5 & 38.4 \\
\midrule
\multicolumn{15}{@{}l}{\textit{Fact verification on TabFact (accuracy)}} \\
\quad original request & -- & 89.7 & -- & 85.0 & -- & 88.8 & -- & 78.7 & -- & 79.0 & -- & 82.5 & -- & 80.5 \\
\quad identical re-send & 1.9 & 89.5 & 0.4 & 85.2 & 0.1 & 88.8 & 0.0 & 78.7 & 13.5 & 75.6 & 0.7 & 82.6 & 0.7 & 81.0 \\
\quad options in reverse order & 2.4 & 89.4 & 0.4 & 85.2 & 4.0 & 89.3 & 12.3 & 74.9 & 14.4 & 78.1 & 5.8 & 83.9 & 4.7 & 79.9 \\
\quad option names A, B, \dots & 2.2 & 89.3 & 2.1 & 85.2 & 2.1 & 88.9 & 9.6 & 79.5 & 21.3 & 72.0 & 3.2 & 83.1 & 4.2 & 81.2 \\
\quad option names replaced by synonyms & 2.1 & 89.3 & 3.6 & 85.2 & 2.4 & 89.5 & 6.8 & 80.0 & 47.6 & 47.0 & 5.1 & 83.7 & 5.8 & 81.5 \\
\quad no option descriptions & 2.2 & 89.5 & 3.0 & 85.8 & 4.3 & 88.5 & 7.6 & 77.0 & 13.7 & 77.8 & 5.3 & 82.5 & 4.2 & 80.2 \\
\quad yes/no question & 4.1 & 89.2 & 4.0 & 86.2 & 4.0 & 89.2 & 12.0 & 79.2 & 13.6 & 80.7 & 6.9 & 81.8 & 5.8 & 80.5 \\
\quad paraphrased instructions & 2.1 & 89.1 & 2.7 & 86.5 & 2.5 & 88.6 & 8.6 & 78.8 & 10.9 & 80.2 & 5.1 & 81.2 & 4.7 & 79.8 \\
\cmidrule(l){1-15}
\quad unverified hint appended & 4.1 & 87.7 & 3.5 & 85.0 & 4.8 & 89.2 & 8.3 & 78.1 & 11.1 & 80.5 & 6.5 & 80.7 & 11.6 & 76.4 \\
\quad no task description & 4.2 & 88.4 & 4.5 & 85.4 & 3.2 & 88.1 & 9.5 & 78.5 & 16.8 & 74.4 & 5.5 & 81.0 & 5.9 & 79.9 \\
\midrule
\multicolumn{15}{@{}l}{\textit{Anomaly detection on ReTabAD (AUROC)}} \\
\quad original request & -- & 74.6 & -- & 74.7 & -- & 81.2 & -- & 80.5 & -- & 77.3 & -- & 71.0 & -- & 61.0 \\
\quad identical re-send & 2.8 & 74.3 & 1.2 & 74.8 & 0.1 & 81.2 & 0.0 & 80.5 & 13.3 & 76.3 & 2.8 & 71.5 & 1.6 & 60.2 \\
\quad options in reverse order & 3.4 & 74.8 & 1.2 & 74.8 & 2.5 & 81.1 & 21.7 & 80.0 & -- & -- & -- & -- & -- & -- \\
\quad option names A, B, \dots & 4.2 & 74.9 & 7.8 & 76.3 & 3.3 & 81.4 & 16.5 & 79.7 & -- & -- & -- & -- & -- & -- \\
\quad option names replaced by synonyms & 3.5 & 73.4 & 5.4 & 74.4 & 2.2 & 80.6 & 6.4 & 81.7 & -- & -- & -- & -- & -- & -- \\
\quad no option descriptions & 9.5 & 71.8 & 8.4 & 73.4 & 7.1 & 78.2 & 13.9 & 76.2 & -- & -- & -- & -- & -- & -- \\
\quad original, binary datasets & -- & 74.6 & -- & 74.7 & -- & 81.2 & -- & 80.6 & -- & -- & -- & -- & -- & -- \\
\quad yes/no question & 6.4 & 73.6 & 10.9 & 74.2 & 4.7 & 79.2 & 21.8 & 80.8 & -- & -- & -- & -- & -- & -- \\
\quad paraphrased instructions & 4.2 & 74.8 & 5.1 & 74.2 & 3.9 & 81.1 & 10.4 & 78.4 & 16.1 & 73.1 & 5.9 & 70.5 & 14.7 & 62.6 \\
\cmidrule(l){1-15}
\quad unverified hint appended & 15.6 & 72.4 & 8.6 & 71.7 & 10.6 & 76.3 & 16.4 & 74.4 & 16.6 & 76.9 & 10.9 & 67.8 & 36.3 & 55.5 \\
\quad no task description & 5.8 & 72.9 & 5.1 & 74.3 & 5.1 & 81.4 & 10.2 & 77.3 & 16.8 & 74.5 & 6.1 & 71.8 & 7.4 & 61.5 \\
\bottomrule
\end{tabular}
\caption{How the question is asked, raw results. \emph{Flip}: share of items whose decision differs from the model's own decision under the original request (\%; invalid outputs count as flips). \emph{Score}: the task metric $\times$100 (row classification: Macro-F1 averaged over the 10 datasets; TabFact: accuracy; anomaly detection: AUROC averaged over the 20 sources). \emph{Identical re-send}: the original request sent again, which gives the re-send rate. Conditions below the line remove or mislead information. Clef sorts the options, so reversing them cannot reach it. On row classification the LLMs receive no option descriptions, and on anomaly detection no options. Llama-3.3-70B answers the yes/no question mostly with ``Yes'' or ``Yes.'', which the strict parse counts as invalid; with a parse that ignores case and a final full stop its flip rate there is 13.0\%. Invalid outputs above 5\% (\%): Llama-3.3-70B on row classification, yes/no question: 54.8; GPT-6 Luna on TabFact, option names A, B, \dots: 9.8; GPT-6 Luna on TabFact, option names replaced by synonyms: 43.7; GPT-6 Luna on TabFact, no task description: 6.6; GPT-6 Luna on ReTabAD, paraphrased instructions: 7.0; Llama-3.3-70B on ReTabAD, paraphrased instructions: 9.5; GPT-6 Luna on ReTabAD, no task description: 5.2. --: does not apply.}
\label{tab:raw_r1}
\end{table*}

\FloatBarrier

\begin{table*}[t]
\centering\scriptsize
\setlength{\tabcolsep}{2.4pt}
\begin{tabular}{@{}lrrrrrrrrrrrrrr@{}}
\toprule
& \multicolumn{8}{c}{\textit{decision models}} & \multicolumn{6}{c}{\textit{LLMs}} \\
\cmidrule(lr){2-9}\cmidrule(l){10-15}
& \multicolumn{2}{c}{\jev{}} & \multicolumn{2}{c}{Clef} & \multicolumn{2}{c}{Kev-27B} & \multicolumn{2}{c}{\shortstack[l]{GPT-6 Luna\\Decisions}} & \multicolumn{2}{c}{\luna{}} & \multicolumn{2}{c}{Qwen3.8-27B} & \multicolumn{2}{c}{Llama-3.3-70B} \\
\cmidrule(lr){2-3}\cmidrule(lr){4-5}\cmidrule(lr){6-7}\cmidrule(lr){8-9}\cmidrule(lr){10-11}\cmidrule(lr){12-13}\cmidrule(lr){14-15}
Condition & Flip & Score & Flip & Score & Flip & Score & Flip & Score & Flip & Score & Flip & Score & Flip & Score \\
\midrule
\multicolumn{15}{@{}l}{\textit{Row classification (Macro-F1)}} \\
\quad original request & -- & 52.1 & -- & 58.4 & -- & 55.2 & -- & 61.2 & -- & 62.5 & -- & 52.2 & -- & 44.0 \\
\quad identical re-send & 4.7 & 52.4 & 1.6 & 58.0 & 0.6 & 54.5 & 0.0 & 61.2 & 13.3 & 63.0 & 1.1 & 52.9 & 0.4 & 44.1 \\
\quad columns shuffled & 15.9 & 51.6 & 15.0 & 57.8 & 15.6 & 54.8 & 18.5 & 57.0 & 20.1 & 60.3 & 20.4 & 52.5 & 17.2 & 46.3 \\
\quad columns reversed & 17.5 & 52.8 & 17.6 & 59.4 & 15.7 & 54.7 & 19.9 & 57.5 & 23.1 & 62.7 & 15.2 & 55.7 & 15.1 & 40.7 \\
\quad questions shuffled within a request & 4.8 & 51.9 & 14.9 & 57.5 & 0.4 & 55.3 & 0.0 & 61.2 & -- & -- & -- & -- & -- & -- \\
\midrule
\multicolumn{15}{@{}l}{\textit{Fact verification on TabFact (accuracy)}} \\
\quad original request & -- & 89.7 & -- & 85.0 & -- & 88.8 & -- & 78.7 & -- & 79.0 & -- & 82.5 & -- & 80.5 \\
\quad identical re-send & 1.9 & 89.5 & 0.4 & 85.2 & 0.1 & 88.8 & 0.0 & 78.7 & 13.5 & 75.6 & 0.7 & 82.6 & 0.7 & 81.0 \\
\quad columns shuffled & 6.3 & 88.2 & 8.0 & 84.2 & 6.2 & 87.9 & 12.7 & 78.3 & 19.9 & 75.4 & 8.2 & 81.7 & 11.0 & 79.4 \\
\quad columns reversed & 7.3 & 87.8 & 9.0 & 85.2 & 6.5 & 87.4 & 13.4 & 77.8 & 20.4 & 74.8 & 9.2 & 80.8 & 12.1 & 78.2 \\
\quad rows shuffled & 6.6 & 88.6 & 7.0 & 85.6 & 6.7 & 88.3 & 11.6 & 77.8 & 18.5 & 75.0 & 8.1 & 82.4 & 9.1 & 80.0 \\
\quad rows reversed & 6.0 & 88.9 & 8.3 & 85.5 & 7.3 & 88.5 & 12.9 & 78.1 & 20.4 & 74.2 & 8.0 & 82.5 & 10.4 & 78.5 \\
\quad table transposed & 8.7 & 86.4 & 9.7 & 83.4 & 8.0 & 86.7 & 17.0 & 73.9 & 25.5 & 70.9 & 12.1 & 77.0 & 16.1 & 74.1 \\
\midrule
\multicolumn{15}{@{}l}{\textit{Anomaly detection on ReTabAD (AUROC)}} \\
\quad original request & -- & 74.6 & -- & 74.7 & -- & 81.2 & -- & 80.5 & -- & 77.3 & -- & 71.0 & -- & 61.0 \\
\quad identical re-send & 2.8 & 74.3 & 1.2 & 74.8 & 0.1 & 81.2 & 0.0 & 80.5 & 13.3 & 76.3 & 2.8 & 71.5 & 1.6 & 60.2 \\
\quad columns shuffled & 7.7 & 73.5 & 9.2 & 73.9 & 6.0 & 79.9 & 10.1 & 79.7 & 17.6 & 75.6 & 7.3 & 70.1 & 15.0 & 59.1 \\
\quad columns reversed & 7.6 & 74.0 & 10.6 & 73.9 & 7.1 & 79.8 & 11.8 & 80.9 & 18.4 & 77.1 & 8.4 & 70.4 & 18.7 & 58.5 \\
\quad rows shuffled & 7.6 & 74.0 & 10.2 & 73.7 & 7.1 & 81.5 & 10.8 & 80.0 & 20.1 & 73.7 & 8.9 & 71.6 & 24.7 & 58.8 \\
\quad rows reversed & 8.8 & 74.4 & 10.6 & 72.3 & 7.6 & 81.0 & 10.7 & 80.6 & 20.0 & 79.2 & 11.6 & 72.5 & 27.2 & 58.6 \\
\bottomrule
\end{tabular}
\caption{Order of rows and columns, raw results. \emph{Flip}: share of items whose decision differs from the model's own decision under the original request (\%; invalid outputs count as flips). \emph{Score}: the task metric $\times$100 (row classification: Macro-F1 averaged over the 10 datasets; TabFact: accuracy; anomaly detection: AUROC averaged over the 20 sources). \emph{Identical re-send}: the original request sent again, which gives the re-send rate. Shuffling the questions of a request applies only to the decision models, which answer 16 questions per request; the LLMs receive one row per prompt. Reordering rows can change the truth of TabFact claims with position words. Invalid outputs above 5\% (\%): GPT-6 Luna on TabFact, columns shuffled: 6.7; GPT-6 Luna on TabFact, columns reversed: 7.2; GPT-6 Luna on TabFact, rows shuffled: 5.9; GPT-6 Luna on TabFact, rows reversed: 6.8; GPT-6 Luna on TabFact, table transposed: 8.2; GPT-6 Luna on ReTabAD, columns shuffled: 7.3; GPT-6 Luna on ReTabAD, columns reversed: 7.1; GPT-6 Luna on ReTabAD, rows shuffled: 8.8; GPT-6 Luna on ReTabAD, rows reversed: 8.0. --: does not apply.}
\label{tab:raw_r2}
\end{table*}

\begin{figure*}[t]
\centering
\includegraphics[width=\textwidth]{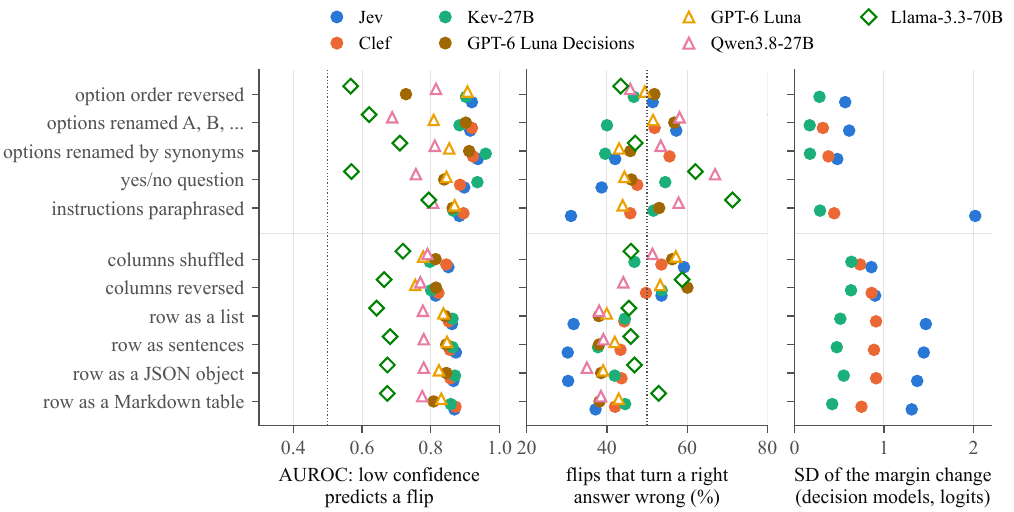}
\caption{Row classification, by rewrite. Left: flip AUROC. Middle: among flips that change
correctness, the share that turn a right answer wrong. Right: standard deviation of the change of the decision models'
margin. Clef is omitted for option order, since it sorts the options.}
\label{fig:byedit}
\end{figure*}
\FloatBarrier

\begin{table*}[t]
\centering\scriptsize
\setlength{\tabcolsep}{2.4pt}
\begin{tabular}{@{}lrrrrrrrrrrrrrr@{}}
\toprule
& \multicolumn{8}{c}{\textit{decision models}} & \multicolumn{6}{c}{\textit{LLMs}} \\
\cmidrule(lr){2-9}\cmidrule(l){10-15}
& \multicolumn{2}{c}{\jev{}} & \multicolumn{2}{c}{Clef} & \multicolumn{2}{c}{Kev-27B} & \multicolumn{2}{c}{\shortstack[l]{GPT-6 Luna\\Decisions}} & \multicolumn{2}{c}{\luna{}} & \multicolumn{2}{c}{Qwen3.8-27B} & \multicolumn{2}{c}{Llama-3.3-70B} \\
\cmidrule(lr){2-3}\cmidrule(lr){4-5}\cmidrule(lr){6-7}\cmidrule(lr){8-9}\cmidrule(lr){10-11}\cmidrule(lr){12-13}\cmidrule(lr){14-15}
Condition & Flip & Score & Flip & Score & Flip & Score & Flip & Score & Flip & Score & Flip & Score & Flip & Score \\
\midrule
\multicolumn{15}{@{}l}{\textit{Row classification (Macro-F1)}} \\
\quad original request & -- & 52.1 & -- & 58.4 & -- & 55.2 & -- & 61.2 & -- & 62.5 & -- & 52.2 & -- & 44.0 \\
\quad identical re-send & 4.7 & 52.4 & 1.6 & 58.0 & 0.6 & 54.5 & 0.0 & 61.2 & 13.3 & 63.0 & 1.1 & 52.9 & 0.4 & 44.1 \\
\quad list (``- name: value'') & 15.0 & 56.0 & 14.0 & 60.3 & 16.1 & 54.6 & 17.2 & 64.7 & 17.7 & 66.5 & 17.9 & 58.4 & 21.3 & 47.9 \\
\quad text (``The name is value.'') & 14.1 & 56.2 & 14.3 & 60.5 & 14.4 & 56.2 & 17.9 & 65.0 & 17.5 & 66.2 & 18.4 & 57.4 & 19.0 & 48.5 \\
\quad JSON object & 14.7 & 56.4 & 13.7 & 59.3 & 17.0 & 56.0 & 17.6 & 64.8 & 17.6 & 66.5 & 17.4 & 59.6 & 17.5 & 47.7 \\
\quad Markdown table & 12.8 & 53.8 & 11.8 & 59.2 & 14.0 & 56.0 & 18.3 & 65.3 & 17.4 & 65.3 & 12.4 & 56.1 & 15.6 & 44.8 \\
\cmidrule(l){1-15}
\quad no field descriptions & 8.5 & 51.7 & 6.4 & 56.3 & 7.2 & 52.7 & 12.3 & 61.2 & 16.0 & 61.6 & 6.8 & 52.8 & 13.6 & 39.6 \\
\quad values without names & 40.5 & 46.8 & 34.3 & 51.3 & 41.4 & 44.7 & 28.8 & 47.6 & 32.0 & 49.4 & 22.6 & 39.7 & 25.8 & 36.4 \\
\quad names attached to other fields' values & 33.6 & 45.7 & 21.1 & 53.0 & 29.0 & 49.9 & 24.8 & 44.9 & 38.5 & 46.2 & 12.2 & 44.1 & 16.7 & 41.7 \\
\midrule
\multicolumn{15}{@{}l}{\textit{Fact verification on TabFact (accuracy)}} \\
\quad original request & -- & 89.7 & -- & 85.0 & -- & 88.8 & -- & 78.7 & -- & 79.0 & -- & 82.5 & -- & 80.5 \\
\quad identical re-send & 1.9 & 89.5 & 0.4 & 85.2 & 0.1 & 88.8 & 0.0 & 78.7 & 13.5 & 75.6 & 0.7 & 82.6 & 0.7 & 81.0 \\
\quad Markdown table & 3.1 & 89.6 & 5.0 & 86.2 & 3.5 & 89.6 & 9.6 & 80.8 & 15.7 & 78.7 & 5.7 & 81.3 & 5.3 & 81.0 \\
\quad CSV & 2.7 & 89.7 & 4.3 & 86.0 & 2.9 & 89.2 & 9.2 & 79.8 & 14.8 & 77.1 & 4.3 & 82.2 & 4.3 & 80.5 \\
\quad JSON records & 4.1 & 90.2 & 5.9 & 86.6 & 4.7 & 89.5 & 10.2 & 80.4 & 19.9 & 75.7 & 7.7 & 81.8 & 7.0 & 81.2 \\
\quad HTML & 3.6 & 90.0 & 5.6 & 87.0 & 4.3 & 89.4 & 9.9 & 80.9 & 15.2 & 78.8 & 6.6 & 82.5 & 6.8 & 80.7 \\
\quad TAPEX linearization & 4.1 & 88.7 & 5.2 & 85.4 & 3.6 & 89.7 & 9.8 & 79.5 & 20.1 & 72.9 & 7.8 & 80.2 & 6.9 & 81.1 \\
\midrule
\multicolumn{15}{@{}l}{\textit{Anomaly detection on ReTabAD (AUROC)}} \\
\quad original request & -- & 74.6 & -- & 74.7 & -- & 81.2 & -- & 80.5 & -- & 77.3 & -- & 71.0 & -- & 61.0 \\
\quad identical re-send & 2.8 & 74.3 & 1.2 & 74.8 & 0.1 & 81.2 & 0.0 & 80.5 & 13.3 & 76.3 & 2.8 & 71.5 & 1.6 & 60.2 \\
\quad JSON object & 3.8 & 74.3 & 4.9 & 75.0 & 3.0 & 81.1 & 8.9 & 81.2 & 18.9 & 76.8 & 5.0 & 72.5 & 7.9 & 60.7 \\
\quad list (``- name: value'') & 4.3 & 74.3 & 5.4 & 74.6 & 3.5 & 81.0 & 9.0 & 80.3 & 17.7 & 75.6 & 4.6 & 71.8 & 11.2 & 59.6 \\
\quad text (``The name is value.'') & 4.2 & 74.4 & 6.0 & 75.0 & 3.9 & 81.3 & 9.6 & 80.2 & 17.7 & 74.8 & 5.4 & 71.8 & 10.6 & 60.0 \\
\quad Markdown table & 7.8 & 73.0 & 8.6 & 76.3 & 6.2 & 81.0 & 11.6 & 79.4 & 16.3 & 76.5 & 7.6 & 71.0 & 17.8 & 59.7 \\
\quad numbers with three decimals & 3.5 & 75.0 & 5.1 & 74.2 & 3.0 & 80.9 & 8.4 & 79.8 & 16.9 & 76.5 & 6.0 & 71.1 & 5.4 & 61.1 \\
\cmidrule(l){1-15}
\quad values without names & 12.1 & 69.8 & 13.9 & 72.1 & 10.9 & 79.7 & 13.4 & 75.5 & 22.8 & 73.5 & 12.5 & 65.3 & 19.1 & 54.6 \\
\bottomrule
\end{tabular}
\caption{Format of the table, raw results. \emph{Flip}: share of items whose decision differs from the model's own decision under the original request (\%; invalid outputs count as flips). \emph{Score}: the task metric $\times$100 (row classification: Macro-F1 averaged over the 10 datasets; TabFact: accuracy; anomaly detection: AUROC averaged over the 20 sources). \emph{Identical re-send}: the original request sent again, which gives the re-send rate. Conditions below the line remove or mislead information. Rendering numbers with three decimals adds precision. Invalid outputs above 5\% (\%): GPT-6 Luna on TabFact, CSV: 5.1; GPT-6 Luna on TabFact, JSON records: 7.3; GPT-6 Luna on TabFact, TAPEX linearization: 8.9; GPT-6 Luna on ReTabAD, JSON object: 9.7; GPT-6 Luna on ReTabAD, list (``- name: value''): 7.7; GPT-6 Luna on ReTabAD, text (``The name is value.''): 9.2; GPT-6 Luna on ReTabAD, Markdown table: 5.6; GPT-6 Luna on ReTabAD, numbers with three decimals: 8.2; GPT-6 Luna on ReTabAD, values without names: 10.9. --: does not apply.}
\label{tab:raw_r4}
\end{table*}

\FloatBarrier

\begin{table*}[t]
\centering\footnotesize
\setlength{\tabcolsep}{2.6pt}
\begin{tabular}{@{}>{\raggedright\arraybackslash}p{4.0cm}rrrrrrrrrrrr@{}}
\toprule
& \multicolumn{2}{c}{\jev{}} & \multicolumn{2}{c}{Clef} & \multicolumn{2}{c}{Kev-27B} & \multicolumn{2}{c}{\shortstack[l]{GPT-6 Luna\\Decisions}} & \multicolumn{4}{c}{\textit{trained on the same examples}} \\
\cmidrule(lr){2-3}\cmidrule(lr){4-5}\cmidrule(lr){6-7}\cmidrule(lr){8-9}\cmidrule(l){10-13}
\textit{Row classification (Macro-F1)} & Score & $\Delta$ & Score & $\Delta$ & Score & $\Delta$ & Score & $\Delta$ & LogReg & LightGBM & Best & TabPFN \\
\midrule
\quad no examples (zero-shot) & 52.1 & -- & 58.4 & -- & 55.1 & -- & 61.2 & -- & -- & -- & -- & -- \\
\quad 4 examples & 60.2 & +8.1 & 60.0 & +1.6 & 59.0 & +3.8 & 59.6 & $-$1.6 & 51.8 & 45.3 & 55.2 & 51.5 \\
\quad 16 examples (the original ones) & 60.0 & +7.9 & 61.7 & +3.2 & 62.7 & +7.6 & 60.4 & $-$0.8 & 58.6 & 51.8 & 58.7 & 59.5 \\
\quad 64 examples$^b$ & 65.3 & +12.2 & 68.2 & +9.0 & 69.3 & +12.2 & 65.4 & +3.3 & 69.2 & 69.1 & 69.4 & 70.8 \\
\quad as many as fit (30--681) & 66.7 & +14.6 & 70.6 & +12.1 & 73.0 & +17.6 & 69.9 & +8.7 & 72.2 & 76.0 & 76.3 & 77.8 \\
\quad whole training split & -- & -- & -- & -- & -- & -- & -- & -- & 74.7 & 80.0 & 80.0 & -- \\
\cmidrule(l){1-13}
\quad 16 examples, classes interleaved & 62.0 & +9.9 & 61.9 & +3.5 & 62.8 & +7.7 & 63.6 & +2.4 & -- & -- & -- & -- \\
\quad 16 examples, written as text & 57.1 & +5.0 & 62.7 & +4.3 & 61.1 & +5.9 & 63.7 & +2.5 & -- & -- & -- & -- \\
\midrule
\textit{TabFact (accuracy)} & Score & $\Delta$ & Score & $\Delta$ & Score & $\Delta$ & Score & $\Delta$ &  &  &  &  \\
\midrule
\quad no examples (zero-shot) & 89.7 & -- & 85.0 & -- & 88.8 & -- & 78.7 & -- & -- & -- & -- & -- \\
\quad 2 examples & 88.4 & $-$1.3 & 86.2 & +1.1 & 89.8 & +1.0 & 77.9 & $-$0.8 & -- & -- & -- & -- \\
\quad 4 examples & 88.0 & $-$1.7 & 86.4 & +1.3 & 89.8 & +0.9 & 78.1 & $-$0.6 & -- & -- & -- & -- \\
\quad 8 examples & 88.1 & $-$1.6 & 86.6 & +1.5 & 89.6 & +0.8 & 77.5 & $-$1.3 & -- & -- & -- & -- \\
\quad as many as fit (68) & 88.8 & $-$1.0 & 87.2 & +2.1 & 89.8 & +1.0 & 76.1 & $-$2.6 & -- & -- & -- & -- \\
\midrule
\textit{Anomaly detection (AUROC)} & Score & $\Delta$ & Score & $\Delta$ & Score & $\Delta$ & Score & $\Delta$ & IForest & OCSVM & $k$NN & Best \\
\midrule
\quad no normal-sample statistics & 65.4 & $-$9.1 & 67.3 & $-$7.3 & 73.8 & $-$7.4 & 74.6 & $-$6.4 & -- & -- & -- & -- \\
\quad statistics only (the original request) & 74.6 & -- & 74.7 & -- & 81.2 & -- & 80.6 & -- & -- & -- & -- & -- \\
\quad statistics and 16 normal samples & 79.0 & +4.4 & 80.5 & +5.9 & 84.2 & +3.0 & 83.6 & +3.0 & 68.1 & 72.2 & 72.3 & 72.3 \\
\quad statistics and as many normal samples as fit & 80.8 & +6.3 & 84.1 & +9.5 & 86.5 & +5.3 & 85.7 & +5.2 & 75.6 & 76.1 & 78.5 & 78.5 \\
\quad all training normals & -- & -- & -- & -- & -- & -- & -- & -- & 75.7 & 75.7 & 78.8 & 78.8 \\
\bottomrule
\end{tabular}
\caption{In-context examples, raw results. Examples are labelled training rows placed in the request (for TabFact, other tables with a claim and its label; for anomaly detection, normal training records after the normal-sample statistics); the sets are nested. \emph{Score}: task metric $\times$100. \emph{$\Delta$}: change of the score from no examples (anomaly detection: from the original request), on the same rows. The other methods are trained on exactly the same examples: logistic regression with cross-validated regularization, LightGBM with few-row settings, and the best of eight trained classifiers (logistic regression, default and tuned; random forest; $k$-nearest neighbours; histogram gradient boosting, LightGBM, CatBoost and XGBoost), the line of Figure~\ref{fig:r3}; on the normal samples, isolation forest, one-class SVM, $k$-nearest neighbours and the best of five detectors. \emph{TabPFN}: TabPFN-3.5, a tabular foundation model that reads the same examples in its context without training (not run on the whole split); TabPFN v2, run the same way, scores 52.3, 57.3, 69.3, 76.4 with 4, 16, 64 and as many examples as fit. $^b$Without communities (about 100 features: 64 examples exceed the budget); the trained methods are averaged over the same nine datasets. The trained anomaly methods cover 19 of the 20 sources (covertype left out).}
\label{tab:raw_r3}
\end{table*}

\FloatBarrier

\begin{table*}[t]
\centering\footnotesize
\setlength{\tabcolsep}{2.6pt}
\begin{tabular}{@{}>{\raggedright\arraybackslash}p{4.0cm}rrrrrr@{}}
\toprule
& \multicolumn{2}{c}{\luna{}} & \multicolumn{2}{c}{Qwen3.8-27B} & \multicolumn{2}{c}{Llama-3.3-70B} \\
\cmidrule(lr){2-3}\cmidrule(lr){4-5}\cmidrule(lr){6-7}
\textit{Row classification (Macro-F1)} & Score & $\Delta$ & Score & $\Delta$ & Score & $\Delta$ \\
\midrule
\quad no examples (zero-shot) & 62.5 & -- & 52.2 & -- & 44.0 & -- \\
\quad 4 examples & 64.9 & +2.4 & 59.2 & +7.0 & 57.6 & +13.5 \\
\quad 16 examples (the original ones) & 64.5 & +2.0 & 59.5 & +7.3 & 56.7 & +12.6 \\
\quad 64 examples$^b$ & 66.2 & +3.0 & 65.4 & +10.4 & 57.4 & +12.5 \\
\quad as many as fit (30--681) & 69.3 & +6.8 & 67.1 & +14.9 & 60.7 & +16.6 \\
\quad whole training split & -- & -- & -- & -- & -- & -- \\
\cmidrule(l){1-7}
\quad 16 examples, classes interleaved & 65.6 & +3.2 & 60.9 & +8.7 & 56.6 & +12.6 \\
\quad 16 examples, written as text & 64.6 & +2.1 & 59.6 & +7.4 & 55.5 & +11.4 \\
\midrule
\textit{TabFact (accuracy)} & Score & $\Delta$ & Score & $\Delta$ & Score & $\Delta$ \\
\midrule
\quad no examples (zero-shot) & 79.0 & -- & 82.5 & -- & 80.5 & -- \\
\quad 2 examples & 79.3 & +0.4 & 82.6 & +0.1 & 79.9 & $-$0.5 \\
\quad 4 examples & 79.5 & +0.6 & 83.5 & +1.1 & 80.7 & +0.2 \\
\quad 8 examples & 80.2 & +1.2 & 83.8 & +1.3 & 79.8 & $-$0.7 \\
\quad as many as fit (68) & 81.5 & +2.5 & 81.0 & $-$1.5 & 77.5 & $-$2.9 \\
\midrule
\textit{Anomaly detection (AUROC)} & Score & $\Delta$ & Score & $\Delta$ & Score & $\Delta$ \\
\midrule
\quad no normal-sample statistics & 70.8 & $-$6.5 & 58.9 & $-$2.2 & 55.4 & $-$5.6 \\
\quad statistics only (the original request) & 77.3 & -- & 63.8 & -- & 61.0 & -- \\
\quad statistics and 16 normal samples & 80.9 & +3.6 & 73.6 & +9.8 & 68.6 & +7.6 \\
\quad statistics and as many normal samples as fit & 82.3 & +5.0 & 77.3 & +13.4 & 72.3 & +11.4 \\
\quad all training normals & -- & -- & -- & -- & -- & -- \\
\bottomrule
\end{tabular}
\caption{In-context examples, raw results of the LLMs; setup, \emph{Score} and \emph{$\Delta$} as in Table~\ref{tab:raw_r3}, where the methods trained on the same examples are listed. $^b$Without communities (about 100 features: 64 examples exceed the budget).}
\label{tab:raw_r3_llm}
\end{table*}

\begin{figure*}[t]
\centering
\includegraphics[width=\textwidth]{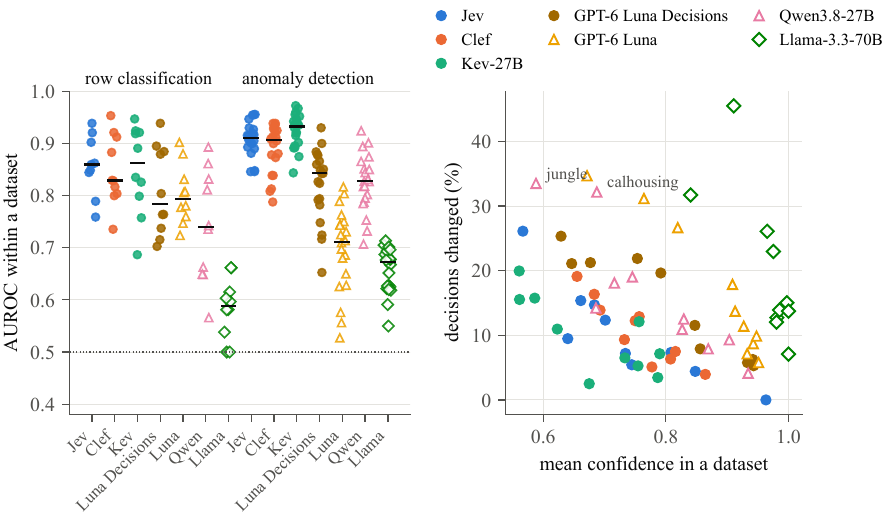}
\caption{Left: flip AUROC within each dataset or source; bars are medians. Right: for each row
classification dataset, the mean confidence of the original decisions against the share of decisions changed by the
information-preserving rewrites.}
\label{fig:datasets}
\end{figure*}
\FloatBarrier

\begin{table*}[t]
\centering\scriptsize
\setlength{\tabcolsep}{2pt}
\begin{tabular}{@{}llrrcccccrrc@{}}
\toprule
& & & & & \multicolumn{4}{c}{\textit{confidence band (1: lowest 25\%)}} & \multicolumn{2}{c}{\textit{noise}} & \\
\cmidrule(lr){6-9}\cmidrule(lr){10-11}
Task & Model & Items & $k$ & All & 1 & 2 & 3 & 4 & Re-send & Bound & Question \\
\midrule
Row classif. & \jev{} & 3,481 & 6 & 33.8\,{\tiny[32.2,35.4]} & 78.0\,{\tiny[75.3,80.7]} & 41.6 & 14.7 & 0.9\,{\tiny[0.3,1.5]} & 4.7 & 25.3 & 20.2\,{\tiny[18.9,21.6]} \\
 & Clef & 3,481 & 6 & 31.7\,{\tiny[30.1,33.2]} & 77.5\,{\tiny[74.8,79.9]} & 37.2 & 10.7 & 1.4\,{\tiny[0.7,2.3]} & 1.6 & 9.0 & 17.3\,{\tiny[16.2,18.5]} \\
 & Kev-27B & 3,481 & 6 & 32.2\,{\tiny[30.6,33.7]} & 76.7\,{\tiny[74.2,79.6]} & 40.0 & 9.0 & 3.1\,{\tiny[2.0,4.3]} & 0.6 & 3.6 & 17.1\,{\tiny[15.8,18.2]} \\
 & \shortstack[l]{GPT-6 Luna\\Decisions} & 3,481 & 6 & 39.6\,{\tiny[38.2,41.2]} & 80.9\,{\tiny[78.3,83.6]} & 52.3 & 23.2 & 1.8\,{\tiny[1.0,2.8]} & 0.0 & 0.0 & 36.7\,{\tiny[35.1,38.2]} \\
\addlinespace[2pt]
 & \luna{}$^a$ & 3,481 & 6 & 42.8\,{\tiny[41.2,44.3]} & 88.7\,{\tiny[86.6,90.9]} & 41.5 & 20.6 & 20.2\,{\tiny[17.6,22.7]} & 13.3 & 57.5 & 34.9\,{\tiny[33.3,36.4]} \\
 & Qwen3.8-27B & 3,481 & 6 & 31.8\,{\tiny[30.4,33.3]} & 61.0\,{\tiny[57.8,64.1]} & 44.6 & 19.0 & 2.6\,{\tiny[1.7,3.7]} & 1.1 & 6.2 & 54.2\,{\tiny[52.4,56.0]} \\
 & Llama-3.3-70B & 3,481 & 6 & 35.5\,{\tiny[34.2,37.2]} & 65.2\,{\tiny[61.9,68.2]} & 27.1 & 23.0 & 26.8\,{\tiny[24.0,29.8]} & 0.4 & 2.4 & 77.1\,{\tiny[75.8,78.6]}$^b$ \\
\midrule
TabFact & \jev{} & 1,755 & 10 & 19.3\,{\tiny[17.5,21.1]} & 54.2\,{\tiny[49.4,58.8]} & 14.8 & 3.6 & 4.6\,{\tiny[2.8,6.4]} & 1.9 & 17.5 & 7.4\,{\tiny[6.1,8.7]} \\
 & Clef & 1,755 & 10 & 23.0\,{\tiny[20.9,25.3]} & 59.9\,{\tiny[54.3,64.6]} & 19.8 & 8.2 & 4.1\,{\tiny[2.4,6.2]} & 0.4 & 3.4 & 8.0\,{\tiny[6.7,9.3]} \\
 & Kev-27B & 1,755 & 10 & 18.3\,{\tiny[16.7,20.2]} & 53.3\,{\tiny[48.7,57.7]} & 13.9 & 3.6 & 2.5\,{\tiny[1.3,4.0]} & 0.1 & 1.5 & 9.8\,{\tiny[8.4,11.2]} \\
 & \shortstack[l]{GPT-6 Luna\\Decisions} & 1,755 & 10 & 38.3\,{\tiny[36.1,40.7]} & 78.1\,{\tiny[74.2,82.2]} & 39.9 & 19.8 & 15.5\,{\tiny[12.3,19.1]} & 0.0 & 0.0 & 26.5\,{\tiny[24.4,28.7]} \\
\addlinespace[2pt]
 & \luna{}$^a$ & 1,749 & 10 & 60.8\,{\tiny[58.2,63.2]} & 88.1\,{\tiny[85.3,91.0]} & 62.0 & 44.2 & 48.7\,{\tiny[43.9,53.6]} & 13.5 & 76.5 & 58.6\,{\tiny[55.9,61.2]} \\
 & Qwen3.8-27B & 1,729 & 10 & 27.4\,{\tiny[25.2,29.7]} & 64.4\,{\tiny[60.0,68.8]} & 23.8 & 11.6 & 9.7\,{\tiny[7.2,12.9]} & 0.7 & 6.8 & 13.9\,{\tiny[12.4,15.6]} \\
 & Llama-3.3-70B & 1,749 & 10 & 33.7\,{\tiny[31.4,36.0]} & 48.6\,{\tiny[44.1,53.0]} & 31.1 & 27.7 & 27.5\,{\tiny[23.2,31.8]} & 0.7 & 6.3 & 13.6\,{\tiny[12.1,15.3]} \\
\midrule
Anomaly det. & \jev{} & 5,059 & 8 & 21.1\,{\tiny[20.0,22.3]} & 62.0\,{\tiny[59.6,64.5]} & 18.1 & 3.6 & 0.6\,{\tiny[0.2,1.1]} & 2.8 & 20.2 & 17.0\,{\tiny[16.0,18.1]} \\
 & Clef & 5,059 & 8 & 25.8\,{\tiny[24.5,27.0]} & 72.8\,{\tiny[70.2,75.4]} & 24.3 & 5.2 & 0.7\,{\tiny[0.3,1.2]} & 1.2 & 9.1 & 21.5\,{\tiny[20.3,22.6]} \\
 & Kev-27B & 5,059 & 8 & 17.7\,{\tiny[16.7,18.8]} & 60.8\,{\tiny[58.0,63.3]} & 9.0 & 0.9 & 0.1\,{\tiny[0.0,0.2]} & 0.1 & 0.8 & 13.0\,{\tiny[12.1,13.9]} \\
 & \shortstack[l]{GPT-6 Luna\\Decisions} & 4,999 & 8 & 29.9\,{\tiny[28.7,31.4]} & 70.3\,{\tiny[68.0,73.0]} & 32.8 & 12.1 & 4.2\,{\tiny[3.3,5.4]} & 0.0 & 0.0 & 38.7\,{\tiny[37.2,40.0]} \\
\addlinespace[2pt]
 & \luna{}$^a$ & 5,014 & 8 & 63.4\,{\tiny[62.0,64.7]} & 85.6\,{\tiny[83.6,87.4]} & 69.5 & 51.0 & 47.6\,{\tiny[45.2,50.1]} & 13.3 & 68.2 & -- \\
 & Qwen3.8-27B & 5,044 & 8 & 23.3\,{\tiny[22.2,24.5]} & 56.9\,{\tiny[54.2,59.7]} & 22.7 & 8.7 & 4.8\,{\tiny[3.7,6.1]} & 2.8 & 20.2 & -- \\
 & Llama-3.3-70B & 4,984 & 8 & 53.4\,{\tiny[52.1,54.8]} & 72.3\,{\tiny[69.6,74.5]} & 50.6 & 44.8 & 45.9\,{\tiny[42.9,48.5]} & 1.6 & 12.1 & -- \\
\bottomrule
\end{tabular}
\caption{Items whose decision depends on how the table is organized (\%). \emph{All}: share of items whose decision changes under at least one of the $k$ information-preserving table rewrites of Table~\ref{tab:rewrites}; \emph{1--4}: the same for the items sorted by the model's confidence in its original decision and split into four bands of 25\%, 1 the lowest (ties broken at random with a fixed seed). \emph{Re-send}: share of decisions changed by an identical re-send of the original request ($p$). \emph{Bound}: $1-(1-p)^k$, the share that would change at least once if every rewrite flipped independently at rate $p$; an upper bound on what noise alone explains, since re-send flips concentrate on the same low-margin items. \emph{Question}: share of items changed by at least one rewrite of the question (options reordered or renamed, yes/no question, option descriptions removed, paraphrased instructions). Brackets (shown for all items, the outer bands and the question): 95\% cluster-bootstrap intervals (Poisson(1) weights, 500 replicates; clusters are classification rows, anomaly-detection samples and TabFact tables). Invalid outputs count as changes. Numbers with three decimals and TabFact claims with position words are left out. Jev's original TabFact decisions carry no probabilities. Confidence: for the decision models, the returned probability of the chosen option; for the LLMs, the share of ten draws at temperature 1 that give the original decision. On anomaly detection the LLMs receive no options, so only one question rewrite (paraphrase) applies to them. $^a$Its decisions are themselves draws at temperature 1, so its counts are dominated by temperature noise. $^b$Mostly answers such as ``Yes.'' to the yes/no question, which the strict parse counts as invalid.}
\label{tab:multiplicity}
\end{table*}

\FloatBarrier

\begin{figure*}[t]
\centering
\includegraphics[width=\textwidth]{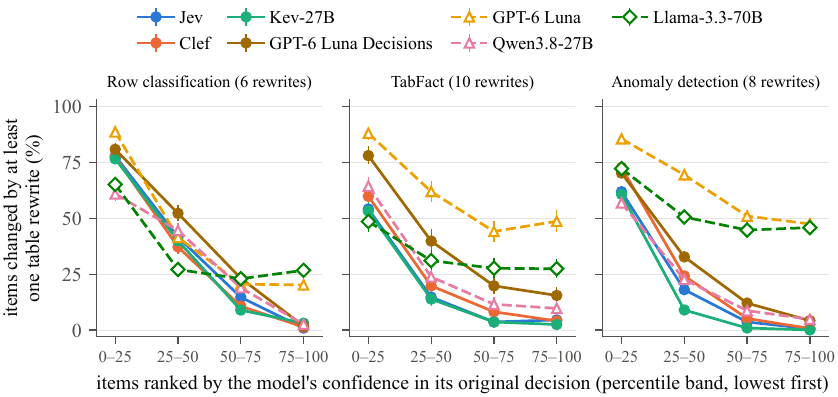}
\caption{Items whose decision changes under at least one information-preserving rewrite of the table. Items are
sorted by the model's confidence in its original decision and split into four bands of 25\%; 95\% cluster-bootstrap
intervals.}
\label{fig:multiplicity}
\end{figure*}

\begin{figure*}[t]
\centering
\includegraphics[width=\textwidth]{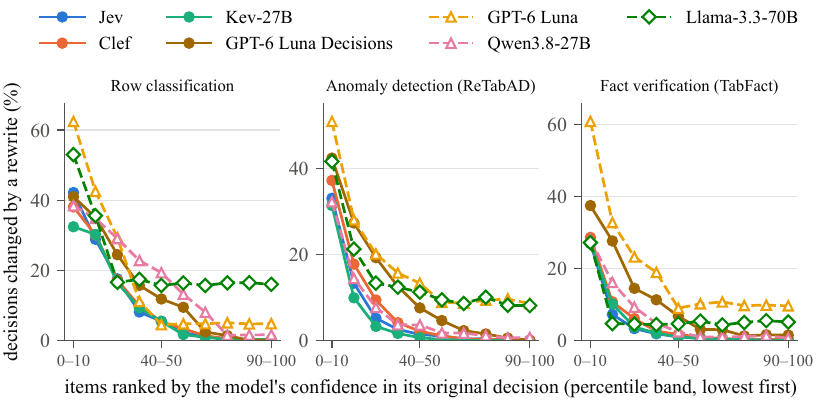}
\caption{Share of decisions changed by a rewrite, for items ranked by the model's confidence in its original decision
and split into bands of 10\% (information-preserving rewrites pooled). Confidence as in the main text: the decision
models' returned probability and the LLMs' share of ten draws that give the original decision; tied values are split
at random.}
\label{fig:confidence}
\end{figure*}
\FloatBarrier

\begin{table*}[t]
\centering\footnotesize
\setlength{\tabcolsep}{5pt}
\begin{tabular}{@{}lccccc@{}}
\toprule
Model & Right (\%), top / low 25\% & stable, right & stable, wrong & fragile, right & fragile, wrong \\
\midrule
\multicolumn{6}{@{}l}{\textit{Row classification}} \\
\quad \jev{} & 67 / 48 & 44 / 0.79 & 23 / 0.78 & 16 / 0.59 & 18 / 0.56 \\
\quad Clef & 86 / 52 & 52 / 0.84 & 16 / 0.78 & 16 / 0.61 & 16 / 0.59 \\
\quad Kev-27B & 85 / 43 & 48 / 0.76 & 19 / 0.67 & 14 / 0.55 & 19 / 0.54 \\
\quad \shortstack[l]{GPT-6 Luna\\Decisions} & 94 / 49 & 49 / 0.91 & 11 / 0.82 & 21 / 0.68 & 19 / 0.62 \\
\addlinespace[2pt]
\quad \luna{} & 82 / 46 & 48 / 0.98 & 9 / 0.95 & 23 / 0.77 & 20 / 0.65 \\
\quad Qwen3.8-27B & 87 / 49 & 51 / 0.88 & 17 / 0.75 & 15 / 0.63 & 17 / 0.62 \\
\quad Llama-3.3-70B & 63 / 54 & 45 / 0.99 & 20 / 0.99 & 17 / 0.91 & 19 / 0.88 \\
\midrule
\multicolumn{6}{@{}l}{\textit{Fact verification on TabFact}} \\
\quad \jev{} & 97 / 73 & 76 / 0.96 & 4 / 0.86 & 14 / 0.78 & 6 / 0.70 \\
\quad Clef & 98 / 60 & 72 / 0.95 & 5 / 0.83 & 14 / 0.79 & 9 / 0.71 \\
\quad Kev-27B & 99 / 69 & 77 / 0.93 & 4 / 0.79 & 12 / 0.73 & 6 / 0.63 \\
\quad \shortstack[l]{GPT-6 Luna\\Decisions} & 97 / 58 & 55 / 0.92 & 7 / 0.78 & 24 / 0.76 & 14 / 0.67 \\
\addlinespace[2pt]
\quad \luna{} & 89 / 61 & 36 / 0.97 & 3 / 0.95 & 43 / 0.86 & 17 / 0.65 \\
\quad Qwen3.8-27B & 94 / 59 & 65 / 0.96 & 7 / 0.81 & 18 / 0.78 & 9 / 0.66 \\
\quad Llama-3.3-70B & 82 / 73 & 58 / 1.00 & 8 / 0.99 & 22 / 0.97 & 11 / 0.92 \\
\midrule
\multicolumn{6}{@{}l}{\textit{Anomaly detection on ReTabAD}} \\
\quad \jev{} & 85 / 58 & 62 / 0.90 & 17 / 0.86 & 12 / 0.66 & 10 / 0.66 \\
\quad Clef & 78 / 57 & 56 / 0.85 & 19 / 0.83 & 14 / 0.63 & 12 / 0.63 \\
\quad Kev-27B & 88 / 52 & 65 / 0.88 & 18 / 0.81 & 9 / 0.61 & 9 / 0.61 \\
\quad \shortstack[l]{GPT-6 Luna\\Decisions} & 88 / 52 & 54 / 0.91 & 16 / 0.85 & 17 / 0.74 & 13 / 0.68 \\
\addlinespace[2pt]
\quad \luna{} & 84 / 65 & 31 / 0.95 & 5 / 0.93 & 46 / 0.84 & 17 / 0.71 \\
\quad Qwen3.8-27B & 85 / 66 & 63 / 0.87 & 14 / 0.84 & 13 / 0.61 & 11 / 0.64 \\
\quad Llama-3.3-70B & 59 / 56 & 29 / 0.95 & 17 / 0.94 & 29 / 0.86 & 25 / 0.85 \\
\bottomrule
\end{tabular}
\caption{Confidence against correctness and fragility (RQ4). Second column: share of right decisions among the 25\% of decisions with the highest and the lowest confidence. Other columns: share of decisions (\%) and their mean confidence in each group; a decision is fragile when at least one rewrite of the table flips it. Confidence as in \S\ref{sec:confidence}.}
\label{tab:confcorrect}
\end{table*}

\end{document}